\documentclass[lettersize,journal]{IEEEtran}
\usepackage{amsmath,amsfonts}
\usepackage{hyperref}
\usepackage{algpseudocode}
\usepackage{algorithm}
\usepackage{arydshln}
\usepackage{array}
\algrenewcommand\algorithmiccomment[1]{\textcolor{gray}{\textit{\# #1}}}
\algrenewcommand\algorithmicrequire{\textbf{Input:}}
\algrenewcommand\algorithmicensure{\textbf{Output:}}

\usepackage[caption=false,font=normalsize,labelfont=rm,textfont=rm]{subfig}
\usepackage{textcomp}
\usepackage{stfloats}
\usepackage{url}
\usepackage{xcolor}
\usepackage{verbatim}
\usepackage{subfig}
\usepackage{graphicx}
\usepackage{cite}
\usepackage{booktabs} 
\usepackage{multirow}
\usepackage{amsmath}     
\usepackage{amssymb} 
\begin{document}

\title{Chain-of-Evidence Multimodal Reasoning for \\Few-shot Temporal Action Localization}

\author{Mengshi Qi,~\IEEEmembership{Member,~IEEE,}
        Hongwei Ji,
        Wulian Yun,
        Xianlin Zhang,
        Huadong Ma,~\IEEEmembership{Fellow,~IEEE} 
\thanks{This work is partly supported by the Funds for the NSFC Project under Grant 62572072 and U24B20176, Beijing Natural Science Foundation (L243027), and Hainan Provincial Joint Project of Li'an International Education Innovation pilot Zone under Grant No.626LALH006. (\emph{Corresponding author: Xianlin Zhang~(email:~zxlin@bupt.edu.cn)})}
\thanks{M. Qi, H. Ji, W. Yun and H. Ma are with State Key Laboratory of Networking and Switching Technology, Beijing University of Posts and Telecommunications, China. X. Zhang is with School of Digital Media and Design Arts, Beijing University of Posts and Telecommunications, China.}
}

\IEEEpubid{}

\maketitle

\begin{abstract}
Traditional temporal action localization (TAL) methods rely on large amounts of detailed annotated data, whereas few-shot TAL reduces this dependence by using only a few training samples to identify unseen action categories. However, existing few-shot TAL methods typically focus solely on video-level information, neglecting textual information, which can provide valuable semantic support for the action localization task.
To address these issues, in this work, we propose a new few-shot temporal action localization method by Chain-of-Evidence multimodal reasoning to improve localization performance. Specifically, we design a novel few-shot learning framework to capture action commonalities and variations, which includes a semantic-aware text-visual alignment module designed to align the query and support videos at different levels.
Meanwhile, to better express the temporal dependencies and causal relationships between actions at the textual level, we design a Chain-of-Evidence (CoE) reasoning method that progressively guides the Vision Language Model (VLM) and Large Language Model (LLM) to generate CoE text descriptions for videos. The generated texts can capture more variance of action than visual features. We conduct extensive experiments on the publicly available ActivityNet1.3, THUMOS14 and our newly collected \emph{Human-related Anomaly Localization Dataset}. The experimental results demonstrate that our proposed method significantly outperforms existing methods in single-instance and multi-instance scenarios. Our source code and data are available at https://github.com/MICLAB-BUPT/VAL-VLM.
\end{abstract}


\section{Introduction}
\IEEEPARstart{W}{ith} the rapid development of social media platforms, such as TikTok and Instagram, the number of short videos has increased tremendously, creating a significant challenge in effectively managing and utilizing large amounts of video resources. Consequently, the importance of video understanding has become more evident.

\begin{figure}
    \centering
    \includegraphics[width=0.8\linewidth]{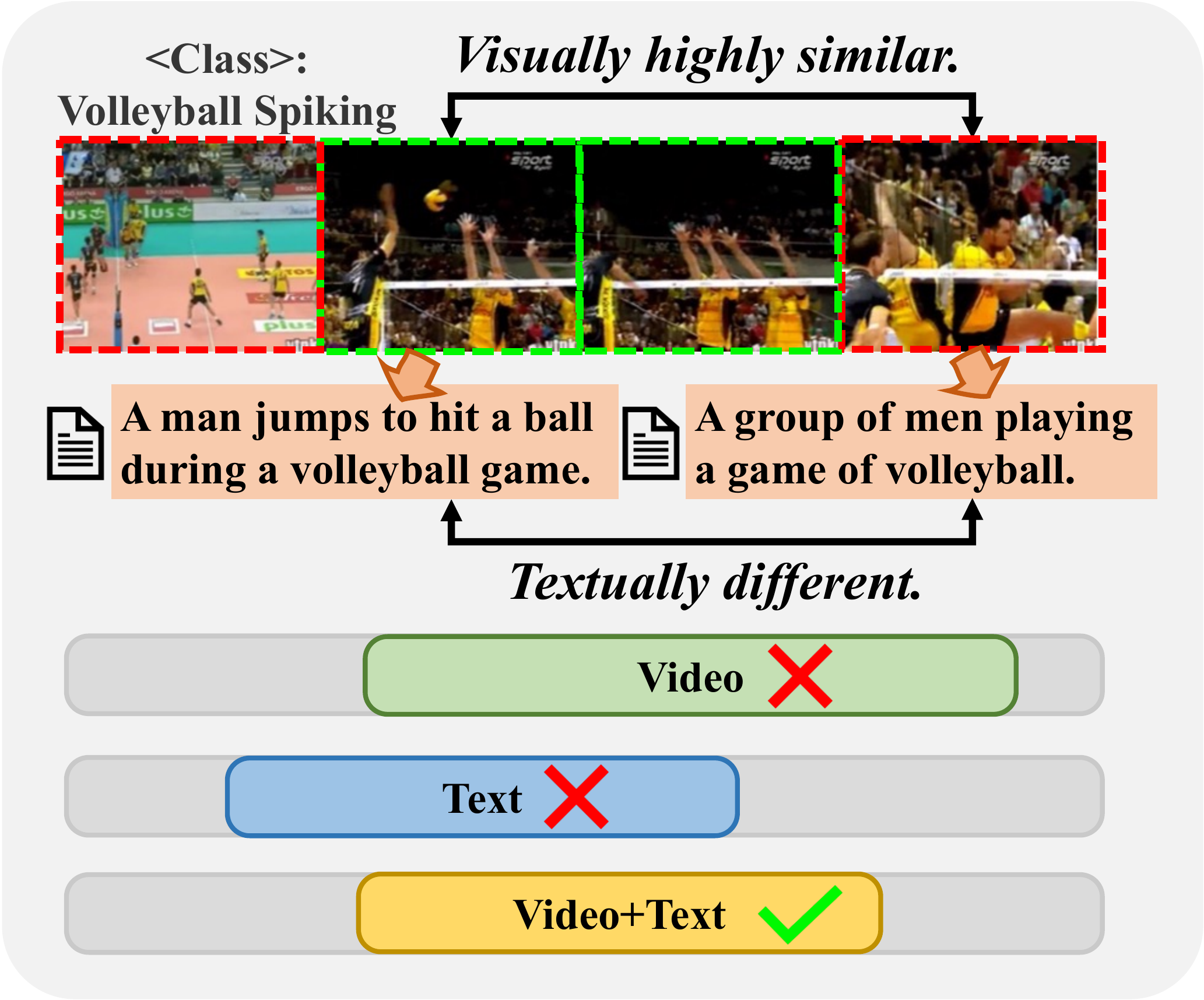}
    % \vspace{-3mm}
    \caption{Illustration of the assistance of multimodal information in few-shot TAL task. Most of existing methods that rely solely on visual information often misjudge when distinguishing between highly similar foreground (\textcolor{green}{green dashed box}) and background (\textcolor{red}{red dashed box}) snippets. The text provides more semantic details, helping the model achieve more precise predictions. } 
    \label{fig:motivation}
    \vspace{-3mm}
\end{figure}

Specially, Temporal Action Localization (TAL)~\cite{7780488,SSN2017ICCV,8100158,Tan_2021_RTD,10.1145/3123266.3123343,9157091,zhang2022actionformer} as a crucial task in video understanding, aims to detect the start and end times of action instances in untrimmed videos.
However, existing fully-supervised TAL methods rely on large amounts of precise temporal annotations for training, 
requiring substantial data for each category, which is both time-consuming and costly. 
Furthermore, these methods can only identify action categories present in the training and lack the ability to predict unseen categories, limiting their practical application.

Nowadays, few-shot learning~\cite{8954109,finn2017model,chen2021meta,8954051,chen2019closer} has shown impressive performance in computer vision, providing a solution to the above challenges. By mimicking the human ability to learn from limited labeled samples, models can quickly adapt to new tasks or categories.
Few-shot learning can be classified into two categories: 
meta-learning~\cite{finn2017model,chen2021meta} and transfer learning~\cite{chen2019closer,dhillon2019baseline}.

Meta-learning enhances the model's ability to quickly adapt to new tasks by training it on multiple tasks, while transfer learning reduces data requirements by transferring knowledge from existing tasks to new ones. Therefore, few-shot learning can be introduced into TAL to localize actions in unseen videos using limited data.
 
Current few-shot TAL methods ~\cite{yang2020localizing,yang2021few,nag2021few,hu2019silco,hsieh2023aggregating,lee2023few} mainly rely on meta-learning, aligning query and support videos to capture commonalities and variations within the same action category. This enables the model to effectively apply learned knowledge to new classes. However, extracting variations and commonalities from a limited number of samples becomes challenging. In contrast, textual information explicitly describes the action’s semantic content and context, helping the model capture its commonalities and variations more effectively. Especially, text descriptions of a short video at various timestamps can bring larger differences than visual appearance. To be specific, recent advancements in pre-trained VLMs~\cite{li2024videochat,internvideo2_5,qwen25vl} offer a new perspective on this issue.
VLMs provide additional prior knowledge by learning joint visual-textual representations from large-scale datasets, particularly in modeling person-object relationships. For example, the semantic information provided by the text generated by the VLMs effectively helps distinguish the athlete's spike action from ordinary volleyball scenes. Relying solely on visual information, the model finds it challenging to differentiate between these two visually similar contents, which leads to difficulties in accurate localization as shown in Figure~\ref{fig:motivation}.
Therefore, how to effectively integrate multimodal information into few-shot TAL tasks, leverage differences in textual representations to overcome the limitations of visual features, enhance the distinction between visually similar content, and improve the consistency between query and support videos remains a challenge.

Furthermore, several methods~\cite{10656995,sharegpt4video} provide frame-level video descriptions while ignoring that the occurrence of actions’ temporal dependencies and causal relationships. For example, after a player catches the basketball, the next likely action is to shoot. It is still difficult for the model to accurately identify such action sequences and their underlying connections. Therefore, generating text that effectively expresses these dependencies and causal relationships to guide few-shot TAL task, thereby enhancing the model's understanding of dynamic relationships between actions.

To address the above-mentioned issues, we propose a novel few-shot TAL method, which utilizes textual semantic information to assist the model in capturing both commonalities and variations within the same class. First, we propose a Chain-of-Evidence (CoE) reasoning method, which hierarchically guides VLM and LLM to identify the 
temporal dependencies of actions and the causal relationships between actions, thereby generating structured CoE textual descriptions.
Next, we employ a semantic-temporal pyramid encoder and the CLIP text encoder to extract video and text features across hierarchical levels from the query and support video, and their corresponding text. Subsequently, we design a semantic-aware text-visual alignment module to perform multi-level alignment between videos and texts, leveraging semantic information to capture both commonalities and variations in actions.

Finally, the aligned features are fed into the prediction head to generate action proposals. 
In addition, we explore human-related anomalous events to expand the application scope of few-shot action localization and introduce the first human-related anomaly localization dataset.

Our contributions can be summarized as follows:
\begin{itemize}
\item We introduce a new few-shot learning method for the TAL task, which leverages hierarchical video features with textual semantic information to enhance the alignment of query and support.

\item We design a novel Chain-of-Evidence reasoning method to generate textual descriptions to effectively and explicitly describe temporal dependencies and causal relationships between actions.

\item We collect and annotate the first benchmark for human-related anomaly localization, which totally includes 12 types of anomalies, 1,072 videos and 2,543,000+ frames.

\item We achieve state-of-the-art performance on public benchmarks, attaining improvements of about 4\% on the ActivityNet1.3 dataset and 12\% on the THUMOS14 dataset under the multi-instance 5-shot scenario compared to the other state-of-the-art methods.
\end{itemize}

\section{Related Work}

\noindent
\textbf{Temporal Action Localization~(TAL).} TAL~\cite{7780488,8100158,Lin_2018_ECCV,SSN2017ICCV,PGCN2019ICCV,zhang2022actionformer,Tan_2021_RTD,10.1007/978-3-031-19830-4_29,yun2024weakly} is an essential task in video understanding~\cite{qi2025action,deng2025global,lv2023disentangled,qi2019attentive,qi2021semantics,yun2024semi,qi2019sports,qi2018stagnet,qi2020stc,qi2025robust,wang2025pitn,lv2025t2sg,ye2025safedriverag,zhu2023unsupervised,qi2025dc,qi2026explainable,qi2026towards,lv2026robo}, which aims to locate the start and end times of actions in untrimmed videos. Prevailing fully-supervised methods can be categorized into \emph{two-stage} and \emph{one-stage} approaches.
\emph{Two-stage} methods first generate a set of candidate proposals, which are then classified and refined. This paradigm has evolved from early sliding-window strategies~\cite{7780488} towards boundary-aware networks~\cite{Lin_2018_ECCV}, with subsequent efforts incorporating temporal structural modeling~\cite{SSN2017ICCV} and graph reasoning~\cite{PGCN2019ICCV} to further improve proposal quality.
In contrast, \emph{one-stage} methods directly predict action boundaries and labels in a single pass to improve efficiency. This category spans from anchor-based approaches~\cite{10.1145/3123266.3123343} to recent anchor-free Transformer architectures~\cite{zhang2022actionformer,Tan_2021_RTD,10.1007/978-3-031-19830-4_29}.
However, such methods rely on a large amount of accurate annotations, making them costly and difficult to generalize to unseen classes. Moreover, most of them focus on common daily and sports activities, neglecting critical applications like localizing human-related anomalies.

\noindent
\textbf{Few-shot Learning~(FSL).} FSL aims to enable models to adapt to new categories or tasks from limited training data. It can be divided into two major categories: meta-learning~\cite{8954109,chen2021meta} and transfer-learning~\cite{chen2019closer,dhillon2019baseline}. 

Early works~\cite{yang2020localizing} focused on aligning global representations of query and support videos. Subsequent research improved alignment quality by introducing multi-level feature correspondence~\cite{keisham2023multi}, Query-Adaptive Transformers~\cite{nag2021few}, or cross-correlation attention modules~\cite{lee2023few}.
Concurrently, other studies explored richer context modeling. These approaches extend alignment to the spatial-temporal domain using Transformers~\cite{yang2021few}, leverage bilateral attention for context awareness~\cite{hsieh2023aggregating}, or utilize Context Graph Convolutional Networks~\cite{CGCN2025} to aggregate information.
Beyond the visual-only meta-learning paradigm, prompt-learning has also been explored for few-shot TAL. PLOT-TAL~\cite{fish2025plot} leverages a pre-trained vision-language model via optimizing class-specific prompts over a fixed set of action classes.
Although these methods have achieved significant progress in visual alignment and context modeling, they almost rely on visual cues. This reliance becomes a critical bottleneck in visually ambiguous scenarios and diminishes the accuracy of the localization. To overcome the limitations of visual-only alignment in prior works, we leverage textual semantic information to assist in the query-support alignment, which enhances the model's performance when facing complex cases.

\noindent
\textbf{Large Language Models.}
The emergence of LLMs has a significant impact on Natural Language Processing, demonstrating a superior ability to generalize across unseen tasks via in-context learning. 
Representative works like the GPT series~\cite{achiam2023gpt,brown2020language} and open-source LLaMA variants~\cite{touvron2023llama,grattafiori2024llama3herdmodels} exhibit remarkable performance across domains. Building on this, researchers have extended LLMs to visual tasks. LLaVA~\cite{li2024llava,video-llava} aligns modalities via instruction tuning, while subsequent works like VideoChat~\cite{li2024videochat} and InternVL~\cite{chen2024internvl} further advance temporal understanding and visual encoder scaling.
Recently, DeepSeek-R1~\cite{guo2025deepseek} demonstrated that Chain-of-Thought~(CoT)\cite{wei2022chain} significantly enhances reasoning by decomposing complex problems.

The Chain-of-Thought (CoT)\cite{wei2022chain} paradigm empowers models to decompose complex problems into sequential reasoning steps. 

Therefore, we design a new Chain-of-Evidence~(CoE) reasoning to indirectly incorporate the reasoning capabilities of models like DeepSeek-R1 into the TAL task, especially by adding more logical analysis of evidence through causal reasoning.

\noindent
\textbf{Generative Models.}
Generative models have also been applied to computer vision tasks. Hu~\emph{et~al.}~\cite{hu2026boosting} and Zhou~\emph{et~al.}~\cite{zhou2026high} transfer semantic knowledge via self-distillation to boost HDR reconstruction from degraded SDR images. Yan~\emph{et~al.}~\cite{yan2025efficient} and Ma~\emph{et~al.}~\cite{ma2025unified}  use a diffusion model with frequency priors for low-light image enhancement. Zhao~\emph{et~al.}~\cite{zhao2026alternating} design an alternating exposure control network to achieve a larger dynamic range in real-world environments. The idea of leveraging generative priors to enhance feature representations is useful. In this work,  we adopt a LLM-based CoE generation pipeline to generate or augment textual descriptions.

\begin{figure*}[t]
    \centering
    \includegraphics[width=0.9\linewidth]{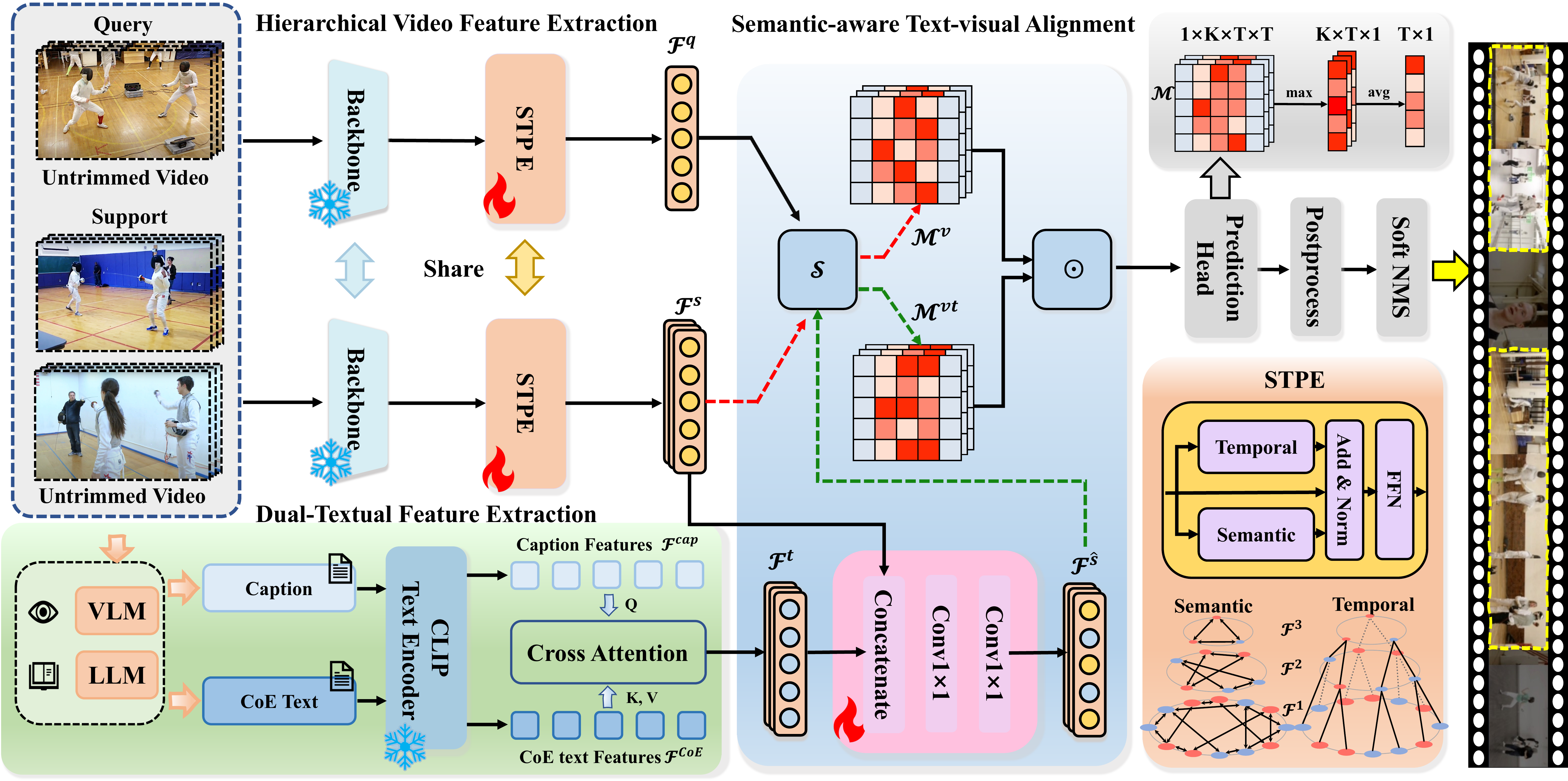}
    \vspace{-3mm}
    \caption{Overall framework of our proposed method.
    We take three inputs: query video, support videos, and the corresponding support text.
    First, the query video and support videos are processed through a pre-trained backbone, followed by feature extraction by the semantic-temporal pyramid encoder (STPE). Next, the textual features of the support text are extracted using the CLIP Text Encoder. Then, the features from the captions and CoE text undergo cross-attention to obtain enhanced textual features. Subsequently, the query video features, support video features, and enhanced textual features are input into the semantic-aware text-visual alignment module. Finally, the aligned features are passed to the prediction head to generate action proposals.}
    \label{fig:framework}
    \vspace{-3mm}
\end{figure*}

\section{Proposed Method}
\subsection{Overview}

\textbf{Problem Definition.} For few-shot TAL, given a training set $D_{\text {train }}=\left\{(x, y) \mid x \in \mathcal{X}_{\text {train}}, y \in \mathcal{C}_{\text {train }}\right\}$ and a test set $D_{\text {test}}=\left\{(x, y) \mid x \in \mathcal{X}_{\text {test}}, y \in \mathcal{C}_{\text {test}}\right\}$. 
Specifically, $x$ denotes the input video and $y={(c, t_s, t_e)}$ represents labels, where $c$ denotes the action category, $t_s$ and $t_e$ represent the start and end time of the action instance, respectively. Note that labels of these two sets are disjoint, \emph{i.e.}, $C_{train}\cap C_{test} = \emptyset$. 
Under the $K$-shot setting, each localization task comprises a query video $v^q$ and $K$ support videos $\{v_i^s\}_{i=1}^K$ with frame-level labels. Our objective is to train a model on $D_{\text{train}}$ that generalizes to $D_{\text{test}}$, enabling it to localize actions in the untrimmed query video by utilizing the annotated support videos.

Our proposed architecture is shown in Figure~\ref{fig:framework}, which mainly consists of Hierarchical Video Feature Extraction, Dual-Textual Feature Extraction, and Semantic-Aware Text-Visual Alignment. Finally, aligned multi-modal features are passed to the prediction head for action localization.

\subsection{Hierarchical Video Feature Extraction}
Given an untrimmed query video $v^q$ and a set of untrimmed support videos $\{v_i^s\}_{i=1}^K$, where $K$ denotes the number of support videos, we first extract features from all the videos.
Specifically,  we first follow~\cite{nag2021few,CGCN2025} to divide the video into non-overlapping snippets, and adopt a pre-trained backbone, \emph{i.e.}, C3D~\cite{7410867}, to extract snippet-level features. Then these features are rescaled to a fixed temporal dimension $T$ using linear interpolation. These features are subsequently fed into Semantic-Temporal Pyramid Encoder to capture more robust features from temporal and semantic levels. Finally, we obtain the feature representation $\mathcal{F}^q \in \mathbb{R}^{1\times T \times D}$ and $\mathcal{F}^s \in \mathbb{R}^{K \times T \times D}$ of the query video and support videos, where $T$ denotes the number of snippets and $D$ is the feature dimension.

\textbf{Semantic-Temporal Pyramid Encoder.}
The visual feature primarily focuses on local motion information, neglecting the modeling of long-term temporal dependencies and semantic relationships. As a result, it fails to adequately capture the temporal sequence of actions and their intrinsic connection with the context. To address this, we propose a new semantic-temporal pyramid encoder~(STPE) to enhance the modeling of both semantic features and long-temporal dependencies at hierarchical levels, as shown in Figure \ref{fig:framework}.
Our STPE mainly contains a temporal pyramid block and a semantic pyramid block. 
First, we follow~\cite{liu2022pyraformer} to establish a pyramid structure to obtain feature representations at different scales. Given a video feature $\mathcal{F}^1=\{f^1_1,f^1_2,\ldots,f^1_{T}\}$ generated by C3D, where $f_t^1$, $ (t = 1, 2, \ldots, T)$ denotes the snippet feature,  we sequentially perform several snippet-level convolution operations along the temporal dimension of video features $\mathcal{F}$, we can extract feature sequences at various scales, which can be expressed as:

\begin{equation}
    \mathcal{F}^{k+1}=\{\Theta(f^k_{3j-2},f^k_{3j-1},f^k_{3j})\}_{j=1}^{\lfloor T/3^k\rfloor},
    \label{equation1}
\end{equation}
where $\Theta$ represents the convolution layer with a kernel size of 3 and a stride of 3, and $\mathcal{F}^{k+1}\in \mathbb{R}^{\lfloor\frac{T}{3^k}\rfloor\times D}(k=1,2,...)$ represents the features after $k$ snippet-level convolution operations.
Subsequently, we stack these feature sequences to form the pyramid structure, as illustrated in Figure \ref{fig:framework}.
For each feature $f^1_t$ in $\mathcal{F}^1$, we compute the temporal attention by considering its adjacent features $\{f^1_{t-1},f^1_{t+1}\}$in the same layer, as well as the feature $f^2_{\lfloor \frac{t+1}{3} \rfloor}=\Theta(f^1_{t-1},f^1_t,f^1_{t+1})$ from the subsequent layer obtained through convolution using the Eq.~(\ref{equation1}). The same attention operation is also performed for other features in the same layer and the higher layers.

However, relying solely on long-term temporal modeling is insufficient for accurately localizing action boundaries, as it fails to capture the intrinsic contextual connections. Therefore, we propose the semantic pyramid block to explore the semantic relationships between snippets. For each feature $f^k_t$ in $\mathcal{F}^k$, we only need to focus on the $m$ most similar features within the same layer to perform a semantic attention operation. 
Specifically, we compute a pairwise cosine similarity matrix $S\in\mathbb{R}^{\lfloor\frac{T}{3^k}\rfloor\times\lfloor\frac{T}{3^k}\rfloor}(k=1,2,...)$ for all features within the layer. For each feature $f^k_t$, we select the $m$ features with the highest similarity scores to form its dynamic semantic nodes $\mathcal{F}^{sim}\in\mathbb{R}^{m\times D}=\{f^k_a,f^k_b,\ldots\}$.
This approach not only helps reduce the computational burden but also enhances the discrimination among features. The top-$m$ similar snippets are selected because snippets from the same action or closely related context tend to have higher feature similarity, while many dissimilar snippets correspond to irrelevant background. This sparse semantic attention also reduces the attention cost from $O(L^2D)$ to $O(LmD)$ for a layer with length $L$. The learning process can be formulated as:
\begin{equation}
    Attn(f^k_t)=softmax\{\frac{f^k_tW_Q}{\sqrt{D}} (\mathcal{F}^{sim} W_K)^T\}\cdot(\mathcal{F}^{sim}W_V),
\end{equation}
where $W_Q, W_K, W_V$ are learnable parameters and $D$ is the feature dimension. Semantic pyramid block enhances the semantic connections across different snippet scales, consolidating commonalities and strengthening variations within the class. Through the two pyramids, a residual connection and a feed-forward neural network are applied.
Finally, we obtain the query video features $\mathcal{F}^q$ and support video features $\mathcal{F}^s$.

\subsection{Dual-Textual Feature Extraction}
\begin{figure*}[t]
    \centering
    \includegraphics[width=0.9\linewidth]{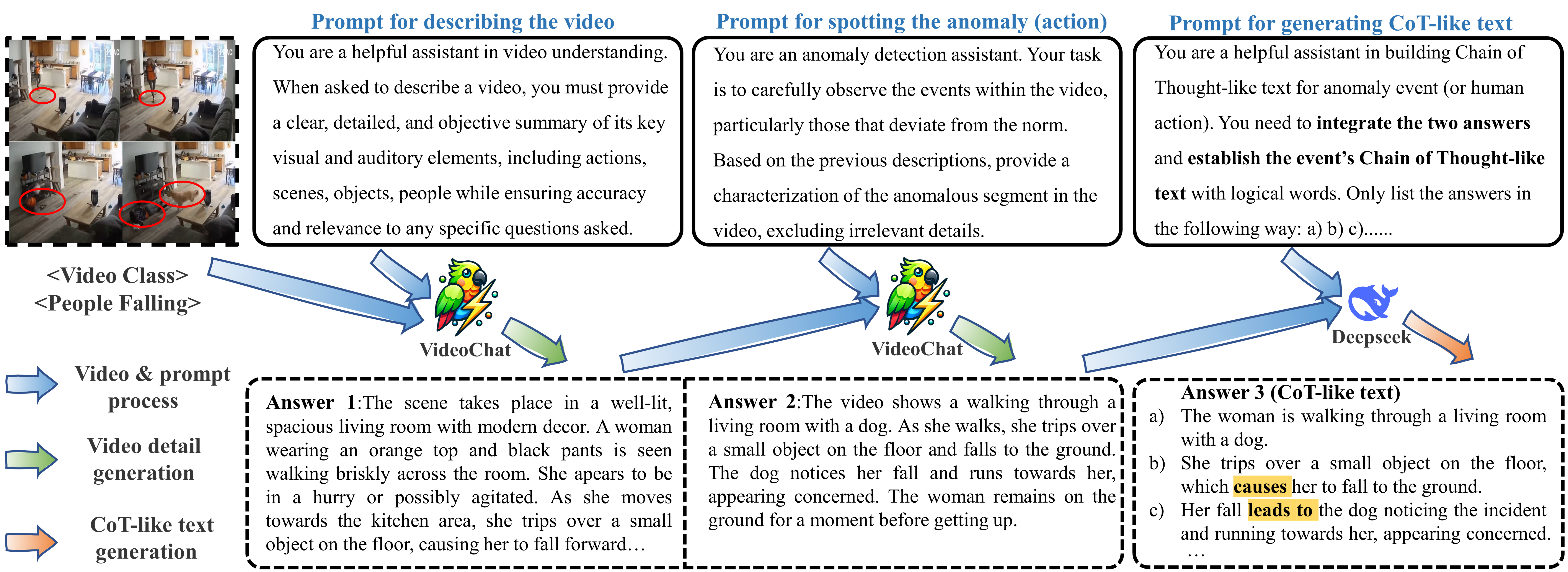}
    \vspace{-2mm}
    \caption{An overview of CoE reasoning. We first prompt (\textbf{\textcolor{blue}{ $\rightarrow$}}) the VLM to generate details of the video and the process of the human action (or anomaly event) for the given video (\textbf{\textcolor{green}{ $\rightarrow$}}). Next, we ask the LLM to generate CoE text based on the details and event sequence provided by the VLM (\textbf{\textcolor{orange}{ $\rightarrow$}}). The logical connectors in the CoE text (Answer 3) have been highlighted in yellow.}
    \vspace{-3mm}
    \label{fig:text_generation}
\end{figure*}

For the video in the support set, we pre-generate the frame-level captions and CoE textual descriptions utilizing the VLM and LLM, as shown in Figure \ref{fig:text_generation}. More details about CoE textual descriptions generation please refer to Section~\ref{subsec:coe}.
Subsequently, the above descriptions are processed through CLIP Text Encoder~\cite{radford2021learning} to extract the corresponding caption features  $\mathcal{F}^{cap}\in\ \mathbb{R}^{1 \times K\times T \times D }$ and CoE text features $\mathcal{F}^{CoE}\in\ \mathbb{R}^{1 \times K\times T' \times D }$. Here, the dimension $T'$ corresponds to the number of sentences in the generated CoE description of the video. The CoE description is generated at the video level rather than at the action-instance level. For each support video, all CoE sentences are encoded in their original order, and no random sentence sampling is used during training. To combine the temporal nature of the caption with the CoE text features, we apply cross-attention between the two to generate the final text feature $\mathcal{F}^t\in\ \mathbb{R}^{1 \times K\times T \times D }$  for assisting the TAL task, which can be formulated as:
\begin{equation}
\mathcal{F}^t=softmax\{\frac{\mathcal{F}^{cap}W_Q}{\sqrt{D}} (\mathcal{F}^{CoE} W_K)^T\}\cdot(\mathcal{F}^{CoE}W_V),
\end{equation}
where $W_Q, W_K, W_V$ are learnable parameters and $D$ is the feature dimension.
This approach enables us to effectively combine these two types of features while preserving the temporal sequence of the caption features and introducing greater coherence and comprehensiveness.

\subsection{Semantic-Aware Text-Visual Alignment}
After obtaining the video feature representations of the query and support, as well as the textual features, denoted as $\mathcal{F}^q$, $\mathcal{F}^s$, and $\mathcal{F}^{t}$. We design a new semantic-aware text-visual alignment module. Firstly we align the video features $\mathcal{F}^q$ and $\mathcal{F}^s$ of query and support, where we utilize cosine similarity to measure the degree of alignment between a query-support snippet pair, resulting in the video alignment map $\mathcal{M}^v\in\ \mathbb{R}^{1 \times K\times T \times T }$, formulated as follows:
\begin{equation}
    \mathcal{M}^v=\mathcal{S}(\mathcal{F}^q,\mathcal{F}^s),
\end{equation}
where $\mathcal{S}$ denotes cosine similarity.
However, solely relying on $\mathcal{M}^v$ to align the query and support action snippets may result in inaccurate alignments, particularly when snippet pairs are irrelevant but share highly similar action backgrounds. 
Hence, we introduce textual information that can explicitly describe the action and background context, aiding in the capture of commonalities and variations. 

Then, we align the support text features $\mathcal{F}^t$ with the support video features $\mathcal{F}^s$ to obtain video-text aligned features $\mathcal{F}^{\hat{s}} \in \mathbb{R}^{K \times T \times D}$. We first concatenate the features from two modalities along the feature dimension and apply two 1×1 convolutions for alignment, which are formulated as:  
\begin{equation}
\mathcal{F}^{\hat{s}}=\Theta \left(\sigma\left(\Theta\left(\mathcal{F}^t \oplus \mathcal{F}^s\right)\right)\right),
\end{equation}
where $\Theta$ denotes the convolution operation, $\sigma $ represents the ReLU activation function, and $\oplus$ means the concatenation along the feature dimension. In this way, we align the features from both modalities, enriching the support set and providing additional auxiliary information for the subsequent query and support video alignment.
Subsequently, to align between the video features $\mathcal{F}^q$ of query and video-text aligned features $\mathcal{F}^{\hat{s}}$ of the support, we calculate the video-text alignment map 
$\mathcal{M}^{vt}\in{\mathbb{R}^{1 \times K\times T \times T }}$ in the same manner of $\mathcal{M}^v$: 
\begin{equation}
    \mathcal{M}^{vt}=\mathcal{S}(\mathcal{F}^q,\mathcal{F}^{\hat{s}}).
\end{equation}

Relying solely on the video alignment map $\mathcal{M}^v$ to align the query and support can easily lead to the misalignment of visually similar foreground and background snippets. In contrast, the video-text alignment map $\mathcal{M}^{vt}$ leverages the clarity of textual semantics to reduce such occurrences. Therefore, we perform an element-wise multiplication of the two maps, using the video-text alignment map $\mathcal{M}^{vt}$ to correct the erroneous regions in the video alignment map $\mathcal{M}^v$. Besides, the background snippets often vary significantly across different support samples, so we concentrate on aligning action commonalities within the foreground snippets by applying a background snippets masking operation on the alignment map. The entire process can be formulated as:
\begin{equation}
\mathcal{M}=\mathcal{M}^v\odot \mathcal{M}^{vt}\odot \mathcal{M}^{m},
\end{equation}
where $\mathcal{M}\in{\mathbb{R}^{1 \times K\times T \times T }}$, $\odot$ is the element-wise multiplication and $\mathcal{M}^{m}$ is the background snippet mask matrix. The mask is constructed only from support annotations and does not use any query label. For the $i$-th support video, its snippet-level label is denoted as $y_i^s\in\{0,1\}^{1\times T}$, where 1 indicates foreground/action and 0 indicates background. We construct a support-specific mask $\mathcal{M}_i^m\in\mathbb{R}^{T\times T}$ by repeating $y_i^s$ along the query temporal dimension, as the following:
\begin{equation}
\mathcal{M}_i^m[t_q,t_s]=y_i^s(t_s),\quad t_q,t_s=1,\ldots,T .
\end{equation}
For a $K$-shot support set, we stack the $K$ support-specific masks and obtain $\mathcal{M}^m\in\mathbb{R}^{1\times K\times T\times T}$, which has the same shape as $\mathcal{M}^v$ and $\mathcal{M}^{vt}$. Thus, the background columns from support snippets are suppressed while the foreground columns are retained. Finally, we utilize a prediction head to obtain the snippet-level predicted result denoted as  $\hat{p}\in\mathbb{R}^{1\times T}$.

\subsection{Optimization and Inference}
\textbf{Loss function.} To optimize our proposed model, we follow~\cite{loss_function} to employ the cross-entropy loss, which consists of two snippet-level losses, \emph{i.e.,} $\mathcal{L}_{fg}$ and $\mathcal{L}_{bg}$. The total loss function $\mathcal{L}$ can be defined as:
\begin{equation}
    \mathcal{L} = \mathcal{L}_{fg}+\mathcal{L}_{bg}.
\end{equation} 
For better classifying the foreground snippet when there are only a few foreground or background snippets present in a query video during the training, we introduce $k_{fg}$ and $k_{bg}$ to deal with the unbalanced issue as follows:
\begin{equation}
\begin{aligned}
k_{fg}  = \min (t,\frac{t}{t_{fg}+\varepsilon} ), \\
k_{bg}   = \min (t,\frac{t}{t_{bg}+\varepsilon} ),
\end{aligned}
\end{equation}
where $t$, $t_{fg}$ and $t_{bg}$ are the number of total snippets, foreground snippets, and background snippets, respectively. Additionally, minimum operation and $\varepsilon$ are used to avoid situations with excessively large $k$ and where the divisor is zero. With the adjustment ratios $k_{fg}$ and $k_{bg}$, the two snippet-level loss functions can be described as:
\begin{equation}
\begin{aligned}
    \mathcal{L}_{fg} &= -k_{fg}\displaystyle\sum_{i=1}^{T}y(i)\log[\hat{p}(i)], \\ 
    \mathcal{L}_{bg} &= -k_{bg}\displaystyle\sum_{i=1}^{T}[1-y(i)]\log[1-\hat{p}(i)],
\end{aligned}
\end{equation}
where $y$ is the query ground truth mask and $\hat{p}\in\mathbb{R}^{1\times T}$ represents the snippet-level prediction.

\textbf{Inference.} During inference, we randomly select a novel class from $C_{test}$, which has never been seen before. For each selected class, we choose 1+$k$ videos as query and support to form a $k$-shot localization task, along with the action snippet annotations of the support. 
For each query video, we generate the foreground probability of each snippet by applying the frozen model. Subsequently, we select the consecutive snippets as proposals where the foreground probability exceeds a predefined threshold. Additionally, we filter out the too-short proposals and calculate the average probability as confidence for the remaining proposals. We then refine the proposals using soft non-maximum suppression~\cite{softnms} with a threshold of 0.7.

\begin{figure*}[t]
    \centering
    \subfloat[]{
        \includegraphics[width=0.33\linewidth]{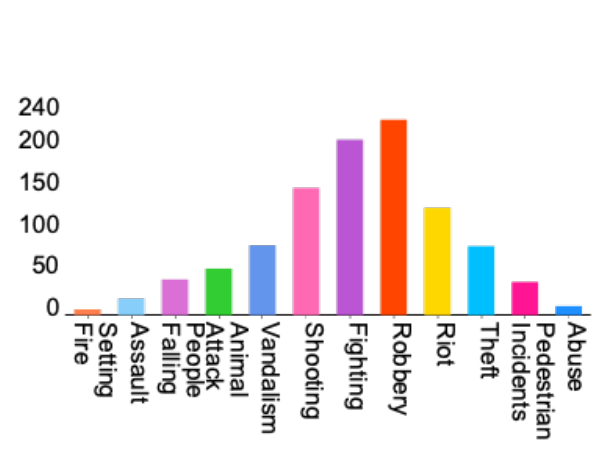}
        \label{fig:dataset_2}
    }
    \subfloat[]{
        \includegraphics[width=0.32\linewidth]{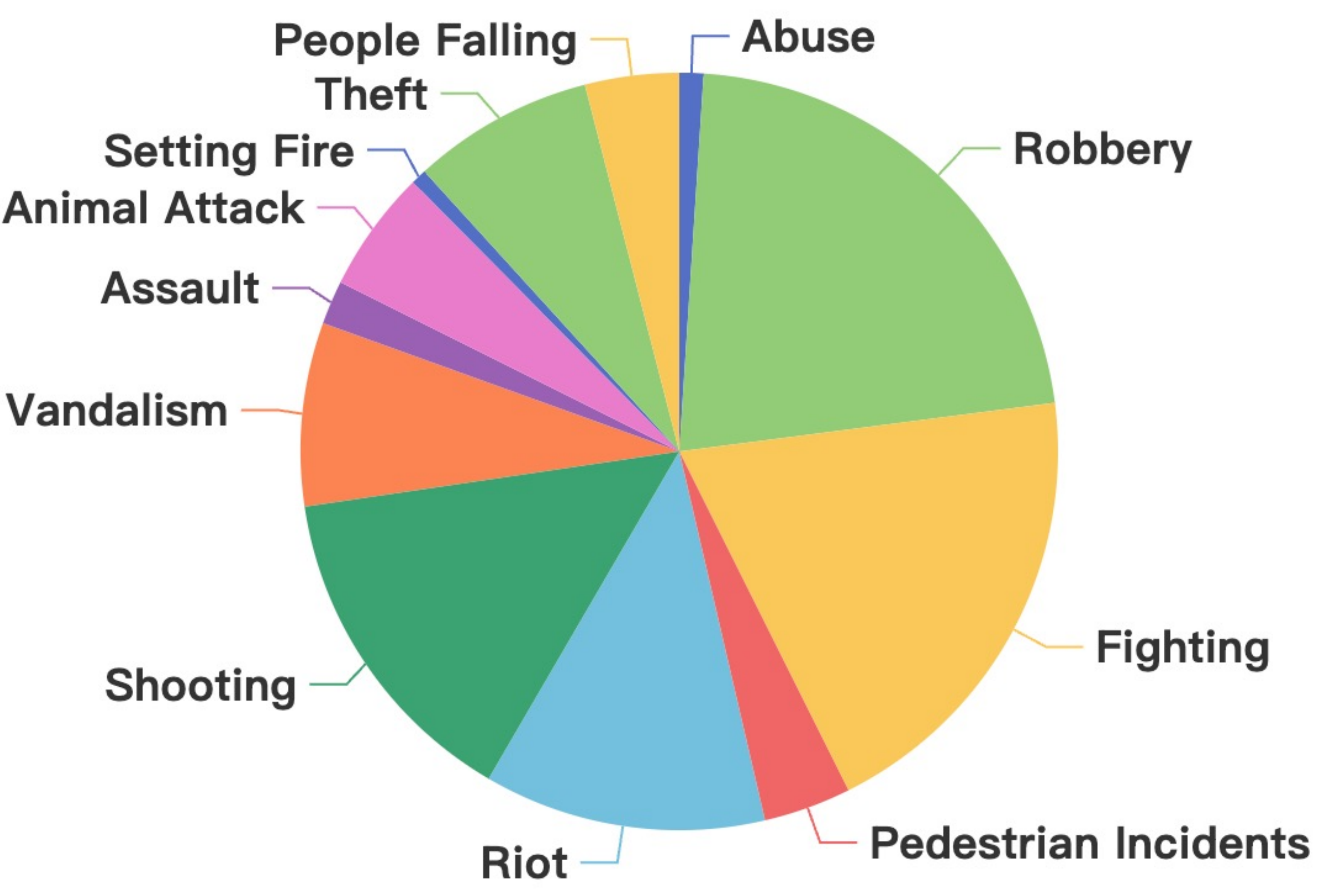}
        \label{fig:dataset_3}
    }
    \subfloat[]{
        \includegraphics[width=0.33\linewidth]{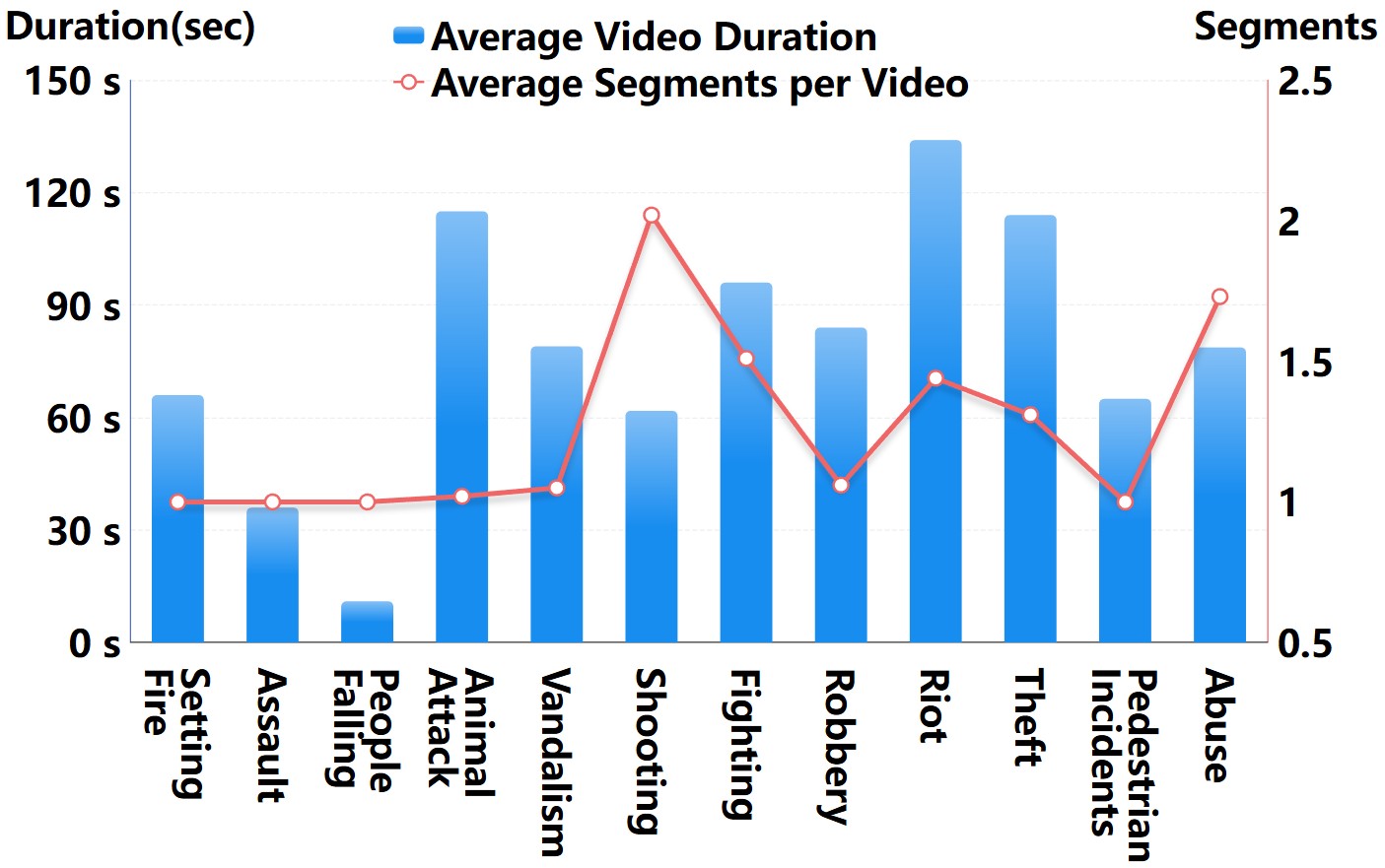}
        \label{fig:dataset_4}
    }
    \caption{
        Statistical overview of the proposed Human-related Anomaly Localization (HAL) dataset. 
        (a) Distribution of the number of videos for each of the 12 anomaly categories.
        (b) Proportions of each anomaly category in the dataset.
        (c) Average duration (in seconds) and the number of distinct anomaly segments per video for each category.
    }
    \label{fig:dataset_statistics}
    \vspace{-3mm}
\end{figure*}

\section{The HAL Benchmark}
To extend the application of temporal action localization to the more practical domains such as human-related anomaly detection, we construct a new Human-related Anomaly Localization (HAL) benchmark. The core feature of HAL is the Chain-of-Evidence (CoE) textual descriptions that we newly generated. Compared to the textual information used in prior works~\cite{10656995,li2023boosting,paul2022text}, this new format is richer in logic and more clearly structured. To efficiently generate the CoE texts, we design an automated CoE reasoning pipeline that guides the VLM and LLM to perform reasoning about the evidence of the causal inference in the video content. The goal is to leverage this causality-infused text to indirectly imbue the localization task with the reasoning capabilities of LLMs, which allows the model to achieve a more precise understanding and localization of complex anomalous events.

\subsection{Data Source}
\begin{figure}
    \centering
    \includegraphics[width=1.0\linewidth]{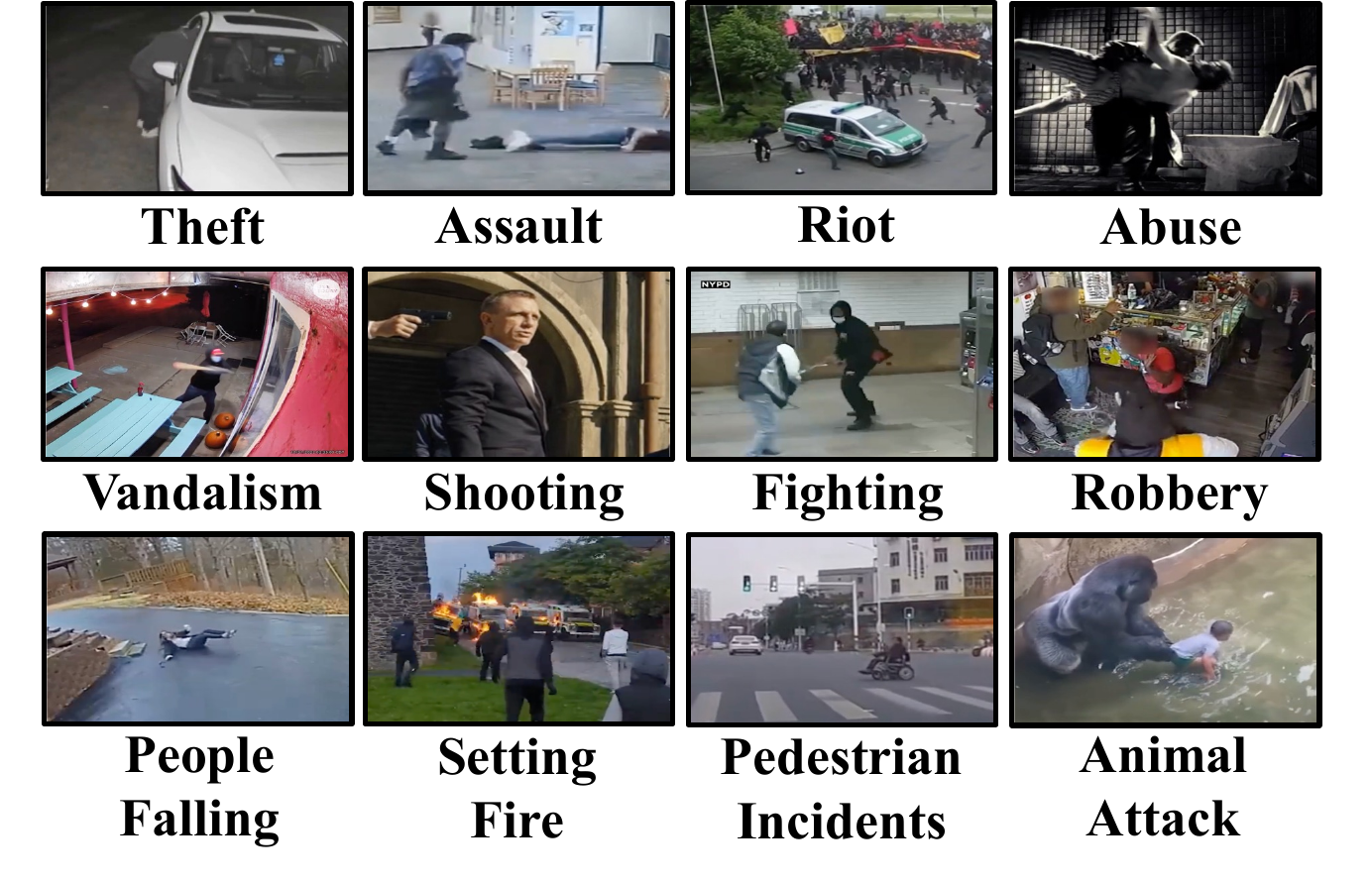}
    \vspace{-8mm}
    \caption{Illustration of 12 anomaly types in our collected HAL dataset.}
    \label{fig:dataset_type}
    \vspace{-2mm}
\end{figure}

Current TAL datasets~\cite{7298698,THU,epic-kitchens} primarily focus on identifying sports and daily activities. However, the task of localizing human anomalous activities is more significant in practical applications, which is of vital importance to the safety of people's lives and property. Hence, we manually select anomalous videos related to human activities from three large-scale anomaly datasets, namely MSAD~\cite{msad2024}, XD-Violence~\cite{XD_Violence}, and CUVA~\cite{CUVA}, and construct the Human-related Anomaly Localization dataset. This dataset contains 12 types of human-related anomalous behaviors, such as fighting, people falling, and robbery, as shown in Figure~\ref{fig:dataset_statistics} and~\ref{fig:dataset_type}. In total, there are 1,072 videos with a cumulative duration of 26.3 hours, comprising over 2,543,000 frames. Each video is accompanied by frame-level annotations of anomaly intervals, along with corresponding frame captions and CoE text.

\begin{figure}
    \centering
    \includegraphics[width=0.95\linewidth]{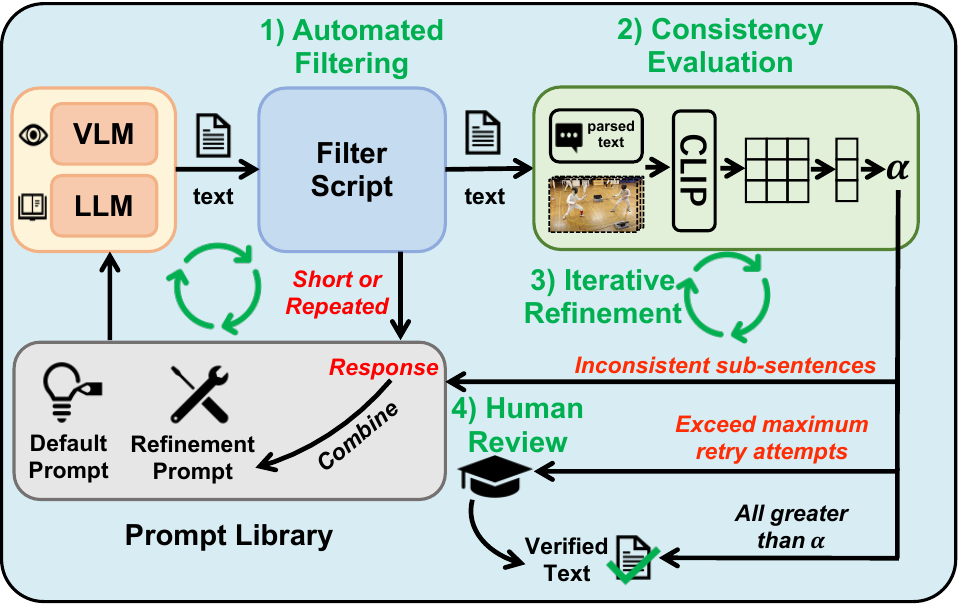}
    \vspace{-2mm}
    \caption{The validation pipeline for the generated text.}
    \label{fig:verify}
    \vspace{-3mm}
\end{figure}

\subsection{Chain-of-Evidence Reasoning Pipeline}
\label{subsec:coe}
To generate texts that adequately represent the temporal dependencies and causal relationships between actions, we propose a new CoE reasoning method, as shown in Figure \ref{fig:text_generation}. 
Our method employs a hierarchical, stage-wise process to guide the VLM and LLM in progressively generating structured, CoE textual descriptions, with each stage refining the prior output to enhance action understanding. 

Specifically, the reasoning process is explicitly decomposed into three progressive stages. Each stage $s_t=(p_t, x_{t-1}, o_t )$ consists of the prompt $p_t$, the input $x_{t-1}$ from previous steps and the corresponding textual output $o_t$.
First, for each video $x$, we utilize the VLM (\emph{i.e.}, VideoChat~\cite{li2024videochat}) to generate the detailed video descriptions denoted as $s_1=(p_1, x_{0}, o_1 )$. Although these descriptions provide detailed depictions of the entire video, they also introduce substantial noisy information. Therefore, we further guide the VLM to filter out such redundant information, enabling it to focus on the most critical actions or anomalous snippets. This process can be denoted as $s_2=(p_2, x_{1}, o_2)$. Building upon this, we further guide the reasoning LLM, such as DeepSeek-R1~\cite{guo2025deepseek}, to perform in-depth logical analysis and reasoning on the generated video-level descriptions $o_1$ and $o_2$, enabling the identification of action sequences and underlying causal relationships. The subsequent reasoning process can be represented as $s_3=(p_3, x_2, o_3 )$, where $x_2=\{o_1,o_2\}$, $o_3=\{c_1,c_2,\ldots\}$, $c_i$ can be a sub action $e$ from $\{o_1,o_2\}$ or a causal link $e_i\rightarrow e_{i+1}$. Taking Figure \ref{fig:text_generation} as an example, the first element $c_1$ in $o_3$ describes a standalone action ($e_1$): ``The woman is walking through a living room with a dog.'' The following elements then capture causal links $e_i \rightarrow e_{i+1}$, explicitly stating that ``She trips over a small object... ($e_1$) which \textbf{causes} her to fall ($e_2$)'' and ``Her fall ($e_2$) \textbf{leads to} the dog noticing the incident... ($e_3$)'' thereby constructing chains of evidence.

To ensure the coherence and logical integrity of the generated chain, the prompt $p_3$
 is specifically engineered to instruct the LLM to maintain consistency with the entities, scenes, and temporal flow established in the previous stages. By explicitly tasking the model with identifying and connecting key events, the prompt activates the inherent reasoning capabilities to construct a coherent narrative, where each step logically follows from the last. This prevents disconnected or contradictory statements and ensures a chain-like structure in the final output $o_3$.
Through this multi-stage generation process, the resulting text progressively presents a structured CoE description. Finally, we generate approximately 7,000, 87,000, and 2,400 CoE sentences/evidence steps for the  HAL, ActivityNet1.3~\cite{7298698} and THUMOS14~\cite{THU}, respectively.

\subsection{Generated Texts Verification Pipeline}
To ensure the quality of our generated CoE texts, we designed a validation pipeline for the output of each stage during Chain of Evidence reasoning:\\
\textbf{1) Automated Filtering:}
We first use scripts to filter out entries with formatting errors or invalid content in the text, such as overly short or repetitive responses. This step quickly removes many low-quality outputs.\\
\textbf{2) Consistency Evaluation:}
To quantify semantic alignment, we employ a CLIP-based evaluation method.
We first parse it into sub-sentences based on punctuation or logical connectors and sample the video at 1 fps. Subsequently, we utilize the CLIP encoders to extract the textual and visual features. Then we compute a cosine similarity matrix $S\in\mathbb{R}^{N_{sent}\times N_{frame}}$ for the sub-sentence and frame features, where $N_{sent}$ and $N_{frame}$ represent the number of sub-sentences and frames, respectively. For each sub-sentence, a matching score is derived by averaging its Top-3 highest similarities. Sub-sentences falling below a predefined threshold $\alpha$ are identified as inconsistent and returned for refinement.\\
\textbf{3) Iterative Refinement:}
We implement a feedback-driven mechanism where inconsistent sub-sentences are injected into a refined prompt alongside the context to guide the language model to specifically revise the inconsistent part, after which the output is re-evaluated through step 2.\\
\textbf{4) Human Review:}
Cases failing validation after fixed retries (\emph{e.g.}, 5) are routed to human review. To ensure reliability, the threshold $\alpha$ is pre-calibrated via human cross-validation on a seed dataset. 
Through this iterative self-refined pipeline, we significantly alleviate the hallucinations and ensure the quality of the CoE text in our benchmark. The details of the algorithm can be referred to in the appendix.

\section{Experiments}
\label{experiments}
\subsection{Datasets and Evaluation Metrics}

\textbf{ActivityNet1.3 Dataset}~\cite{7298698} covers 200 actions, containing 19,994 untrimmed videos.
Following~\cite{feng2018video}, we split the 200 classes into three subsets without any overlap for training (80\%), validation (10\%) and testing (10\%), respectively. For the single-instance scenario, we adopt videos that contain one action snippet. For multi-instance scenarios, we utilize the original videos after filtering out those that contain more than one class category. Hence, we remove the videos with invalid links, leaving approximately 16,800 videos in the final.

\textbf{THUMOS14 Dataset}~\cite{THU} covers 20 action categories, with 200 validation videos and 213 test videos. We reconstruct the dataset division following the meta-learning strategy~\cite{yang2020localizing}. The ratio of training, validation, and test classes follows the same proportions as in ActivityNet1.3. Due to the scarcity of single-instance videos in the original THUMOS14 data, we divide the multi-instance video into non-overlapping snippets, each of which will be regarded as a new single-instance video. Under the multi-instance setting, we continue to use the initial video from THUMOS14, the same as we did for ActivityNet1.3.

\textbf{Human-related Anomaly Localization~(HAL) Dataset} is our newly-collected dataset, which contains 1,072 videos and 12 types of human-related anomalous behaviors, as shown in Figure~\ref{fig:dataset_statistics} and~\ref{fig:dataset_type}. For the few-shot learning setup, we follow the same class splitting protocol as with ActivityNet1.3, dividing the 12 anomaly classes into training, validation, and test sets. 
As illustrated in the statistical analysis in Figure~\ref{fig:dataset_4}, the anomalous events in our HAL dataset are inherently sparse, with most videos containing only one or two distinct anomaly segments. Therefore, we utilize the original videos for all experiments on the HAL dataset without making a distinction between single-instance and multi-instance scenarios.

\textbf{Evaluation Metrics.} We utilize the mean average precision (mAP) as an evaluation metric following~\cite{lee2023few}, and report mAP at an IoU threshold of 0.5. In addition, we report average mAP over multiple tIoU thresholds for a more comprehensive evaluation, including [0.5:0.1:0.9] on ActivityNet1.3 and $\{0.3,0.4,0.5,0.6,0.7\}$ on THUMOS14.

\subsection{Implementation Details}
We adopt the Adam optimizer~\cite{2014Adam} with the learning rate of 5e-6 for ActivityNet1.3 and 1e-6 for THUMOS14 and HAL, implemented in the PyTorch~\cite{paszke2019pytorch} framework on a NVIDIA A6000 GPU. During video feature extraction, we rescale the video feature sequences to $T=100$ snippets for ActivityNet1.3, $T=256$ for THUMOS14, and $T=512$ for HAL using linear interpolation. For ActivityNet1.3, we train 200 epochs and the batch size is set to 100 with 100 episodes per epoch. For THUMOS14, we train 100 epochs and the batch size is set to 20 with 50 episodes per epoch. For HAL, we train 150 epochs and the batch size is set to 30 with 50 episodes per epoch. The CoE text generation pipeline is conducted offline on the support set before model training. Once the generated texts are cached as CLIP features, they introduce no additional cost during the training or test-time localization of the TAL model. On average, the full offline pipeline takes about 34.3 seconds per support video, measured on 4$\times$NVIDIA RTX A6000 GPUs; a detailed per-stage breakdown is provided in the Supplementary Material.

\subsection{Comparison with State-of-the-Art Methods}
We compare our method with the state-of-the-art methods~\cite{feng2018video,yang2020localizing,yang2021few,nag2021few,hu2019silco,hsieh2023aggregating,lee2023few,CGCN2025} on ActivityNet1.3 and THUMOS14 datasets in Table \ref{tab:main_table}. Additionally, we conduct experiments on the HAL dataset and report in the Table \ref{tab:combined_results} (a).

\begin{table*}[t]
\centering
\small
\caption{
    Comparison with state-of-the-art methods in terms of mAP@0.5 on ActivityNet1.3 and THUMOS14 datasets, under single-instance and multi-instance settings. The best results are highlighted in \textbf{bold}, and the second-best results are \underline{underlined}. Ours$^*$ denotes our text-free variant that removes the entire textual branch and retains only the visual branch of the proposed architecture.
}
\vspace{-2mm}
\label{tab:main_table}
\setlength{\tabcolsep}{4mm} 
\begin{tabular}{lcccccccc}
\toprule
\multirow{3}{*}{Method} & \multicolumn{4}{c}{ActivityNet1.3} & \multicolumn{4}{c}{THUMOS14} \\
\cmidrule(lr){2-5} \cmidrule(lr){6-9}
& \multicolumn{2}{c}{Single-instance} & \multicolumn{2}{c}{Multi-instance} & \multicolumn{2}{c}{Single-instance} & \multicolumn{2}{c}{Multi-instance} \\
\cmidrule(lr){2-3} \cmidrule(lr){4-5} \cmidrule(lr){6-7} \cmidrule(lr){8-9}
& 1-shot & 5-shot & 1-shot & 5-shot & 1-shot & 5-shot & 1-shot & 5-shot \\
\midrule
Video Re-localization~\cite{feng2018video}      & 43.5 & -    & 31.4 & -    & 34.1 & -    & 4.3  & -    \\
Silco~\cite{hu2019silco}        & 41.0 & 45.4 & 29.6 & 38.9 & -    & 42.2 & -    & 6.8  \\
Common Action Localization~\cite{yang2020localizing} & 53.1 & 56.5 & 42.1 & 43.9 & 48.7 & 51.9 & 7.5  & 8.6  \\
Few-shot Transformer~\cite{yang2021few}        & 57.5 & 60.6 & 47.8 & 48.7 & -    & -    & -    & -    \\
QAT~\cite{nag2021few}         & 55.6 & 63.8 & 44.9 & 51.8 & 51.2 & 56.1 & 9.1  & 13.8 \\
ABA~\cite{hsieh2023aggregating} & 60.7 & 61.2 & -    & -    & -    & -    & -    & -    \\
CDC-CA~\cite{lee2023few}         & 63.1 & 67.5 & 49.4 & \underline{54.6} & \underline{55.0} & \underline{60.5} & \underline{10.2} & \underline{16.2} \\
CGCN~\cite{CGCN2025}         & 66.6 & 67.2 & -    & -    & \textbf{58.1} & 60.1 & -    & -    \\
\midrule
Ours$^*$ (visual only) & \underline{67.9} & \underline{68.4} & \underline{51.7} & 53.4 & 52.9 & 56.5 & 8.6 & 13.0 \\
\textbf{Ours} & \textbf{69.6} & \textbf{71.5} & \textbf{54.9} & \textbf{58.7} & 54.1 & \textbf{62.6} & \textbf{14.1} & \textbf{18.2} \\
\bottomrule
\end{tabular}
\end{table*}

\begin{table*}[t]
\color{black}
\centering
\caption{Comparison using the average mAP over tIoU thresholds [0.5:0.1:0.9] on ActivityNet1.3, under the single-instance and multi-instance settings. The best results are in \textbf{bold}.}
\vspace{-2mm}
\label{tab:avgmap_anet_revised}
\setlength{\tabcolsep}{4mm}
\begin{tabular}{llcccccc}
\toprule
\textbf{Setting} & \textbf{Method} & \textbf{0.5} & \textbf{0.6} & \textbf{0.7} & \textbf{0.8} & \textbf{0.9} & \textbf{Mean} \\
\midrule
\multirow{3}{*}{Single-instance (1-shot)}
 & Keisham et al.~\cite{keisham2023multi} & 56.7 & 48.9 & 39.6 & 26.6 & 13.5 & 37.1 \\
 & Zhang et al.~\cite{CGCN2025}            & 66.6 & 56.5 & 45.1 & 32.1 & 11.1 & 42.3 \\
 & \textbf{Ours}                           & \textbf{69.6} & \textbf{60.6} & \textbf{50.1} & \textbf{38.5} & \textbf{19.5} & \textbf{47.7} \\
\midrule
\multirow{3}{*}{Single-instance (5-shot)}
 & Keisham et al.~\cite{keisham2023multi} & 65.2 & 56.7 & 45.7 & 34.6 & 13.6 & 43.2 \\
 & Zhang et al.~\cite{CGCN2025}            & 67.2 & 57.8 & 46.3 & 32.2 & 11.4 & 43.4 \\
 & \textbf{Ours}                           & \textbf{71.5} & \textbf{64.1} & \textbf{50.6} & \textbf{38.5} & \textbf{22.9} & \textbf{49.5} \\
\midrule
\multirow{3}{*}{Multi-instance (1-shot)}
 & Nag et al.~\cite{nag2021few}            & 44.9 & 38.0 & 29.2 & 21.4 & 11.2 & 25.9 \\
 & Keisham et al.~\cite{keisham2023multi} & 51.9 & 44.1 & \textbf{34.6} & \textbf{25.1} & \textbf{12.9} & 30.3 \\
 & \textbf{Ours}                           & \textbf{54.9} & \textbf{44.2} & 33.9 & 22.4 & 10.3 & \textbf{33.1} \\
\midrule
\multirow{3}{*}{Multi-instance (5-shot)}
 & Nag et al.~\cite{nag2021few}            & 51.8 & 42.7 & 32.6 & 23.4 & 11.9 & 30.2 \\
 & Keisham et al.~\cite{keisham2023multi} & 57.1 & 45.4 & \textbf{35.5} & \textbf{25.1} & 11.3 & 31.8 \\
 & \textbf{Ours}                           & \textbf{58.7} & \textbf{46.1} & 35.3 & 24.5 & \textbf{13.6} & \textbf{35.6} \\
\bottomrule
\end{tabular}
\end{table*}

\textbf{ActivityNet1.3}. 
Our method consistently outperforms existing few-shot TAL methods in single-instance and multi-instance scenarios across 1-shot and 5-shot settings. In the multi-instance 5-shot case, it achieves 58.7 mAP@0.5, while in the single-instance 1-shot setting, it reaches 71.5 mAP@0.5. This performance can be attributed to two key factors: First, we extract hierarchical features from temporal and semantic dimensions, allowing better localization of action regions and enhancing semantic relationships. Additionally, incorporating textual information improves the model's ability to capture class variations and commonalities, further enhancing alignment and localization. We further evaluate ActivityNet1.3 with average mAP over tIoU thresholds [0.5:0.1:0.9], as shown in Table~\ref{tab:avgmap_anet_revised}. The results indicate that the improvement of our method is not limited to mAP@0.5 but also holds under stricter localization thresholds.

\textbf{THUMOS14}. As shown in Table \ref{tab:main_table}, our method achieves competitive results across various settings, particularly in the multi-instance 1-shot and 5-shot scenarios, where it improves upon \cite{lee2023few} by 38.2\% and 12.3\%, respectively. In the single-instance 1-shot scenario, mAP@0.5 declines because the single-instance snippets in THUMOS14 are individually extracted from multi-instance videos. Each snippet has an incomplete action and a short duration, which hinders our CoE text from providing comprehensive guidance on action sequences. We also compare with prompt-learning/VL-based few-shot TAL methods following PLOT-TAL's THUMOS14 protocol in Table~\ref{tab:plottal_revised}. In addition to our original 5-shot/5-way setting, we further evaluate the 5-shot/20-way setting. For each class, five samples are selected from the THUMOS14 training set; during training, four samples are used as support and the remaining one is rotated as the query; during testing, query videos are selected from the THUMOS14 test set, while the support set is sampled from the five training samples.Under this 20-way protocol, the PLOT-TAL variants obtain higher average mAP than our method, which mainly reflects a difference in target problem rather than in localization design. PLOT-TAL optimizes prompts via optimal transport over the fixed, known class set of the dataset, whereas our method follows the episodic meta-learning paradigm, which is designed to generalize to novel classes unseen during training. As also noted in~\cite{fish2025plot}, the 20-way and 5-way protocols are not directly comparable for this reason. Under the standard 5-way protocol, our method attains the highest average mAP among all compared methods.

\textbf{HAL}. 
Table~\ref{tab:combined_results} (a) presents our results on our collected HAL dataset. As shown in the table, our method outperforms~\cite{CGCN2025} by 2.9 in mAP@0.5 and by 6.5 in mean mAP under the 5-shot scenario. This improvement can be attributed to the CoE text, which provides a comprehensive logical process for the occurrence of abnormal events, thereby assisting the model in accurately locating the anomalous snippets, where the behavioral logic analysis determines abnormality through temporal patterns, contextual correlations, and scene intentions.

\begin{table}[t]
\color{black}
\centering
\caption{Comparison with PLOT-TAL on THUMOS14 (5-shot), using average mAP over tIoU $\in\{0.3,0.4,0.5,0.6,0.7\}$. ``PL''=prompt learning, ``ML''=meta-learning, ``E2E''=end-to-end.}
\vspace{-2mm}
\label{tab:plottal_revised}
\setlength{\tabcolsep}{2.5mm}
\begin{tabular}{@{}llcc@{}}
\toprule
\textbf{Method} & \textbf{Type} & \textbf{Shot/Way} & \textbf{Avg mAP} \\ \midrule
Common Action Loc.~\cite{yang2020localizing}   & ML            & 5/5  & 22.8 \\
Multi-Level Align.~\cite{keisham2023multi}     & ML            & 5/5  & 31.8 \\
QAT~\cite{nag2021few}                          & ML            & 5/5  & 32.7 \\
CoOp~\cite{zhou2022learning}                   & E2E + PL      & 5/20 & 34.65 \\
PLOT-TAL (CLS)~\cite{fish2025plot}             & E2E + PL      & 5/20 & 38.24 \\
PLOT-TAL (Verbose)~\cite{fish2025plot}         & E2E + PL      & 5/20 & 40.59 \\ \midrule
\textbf{Ours}                                  & ML + CoE Text & 5/5  & 36.72 \\
\textbf{Ours}                                  & ML + CoE Text & 5/20 & 35.43 \\
\bottomrule
\end{tabular}
\end{table}

\subsection{Ablation Study}
\textbf{Impact of Different Components.} We evaluate the impact of different components under a multi-instance scenario on THUMOS14 in Table~\ref{tab:combined_results} (b). We first establish our base model by removing the STPE and all operations involving text, and then gradually add the STPE and textual information to the base. As shown in the table, the results improve with the addition of different modules, demonstrating the effectiveness of the various components proposed in this paper.

\begin{table}[t]
\centering
\caption{Main result on the HAL dataset and ablation study on the THUMOS14 dataset.}
\label{tab:combined_results}
\scriptsize
\setlength{\tabcolsep}{1.0mm} 
\renewcommand{\arraystretch}{1.05} 
\begin{tabular}{@{}p{0.48\columnwidth}cccc@{}}
\toprule
\multirow{2}{*}{\textbf{Method}} & \multicolumn{2}{c}{\textbf{1-shot}} & \multicolumn{2}{c}{\textbf{5-shot}} \\
\cmidrule(lr){2-3} \cmidrule(lr){4-5}
& \textbf{0.5} & \textbf{Mean} & \textbf{0.5} & \textbf{Mean} \\
\toprule
\multicolumn{5}{l}{\textbf{(a) Main Results on HAL dataset}} \\
Base & 5.9 & 2.6 & 14.6 & 8.1 \\
Transformer & 32.4 & 20.4 & 34.6 & 21.2 \\
CGCN & 36.3 & 19.8 & 37.1 & 20.2 \\
Ours (on HAL) & \textbf{38.9} & \textbf{25.2} & \textbf{40.0} & \textbf{26.7} \\
\toprule

\multicolumn{5}{l}{\textbf{(b) Impact of Different Components}} \\
Base & 7.6 & 3.0 & 10.6 & 3.7 \\
Base + STPE & 8.6 & 3.2 & 13.0 & 4.4 \\
Base + Text & 13.0 & 4.2 & 14.6 & 4.9 \\
\toprule

\multicolumn{5}{l}{\textbf{(c) Impact of Semantic-Temporal Pyramid Encoder}} \\

w/o STPE & 13.0 & 4.2 & 14.6 & 4.9 \\
w/o TP & 13.5 & 4.2 & 17.1 & 5.0 \\
w/o SP & 13.7 & 4.8 & 17.6 & 5.5 \\
Transformer & 13.4 & 4.6 & 16.7 & 5.3 \\
\toprule

\multicolumn{5}{l}{\textbf{(d) Impact of Different Textual Content}} \\

Prompt & 12.5 & 3.4 & 14.5 & 5.3 \\
Caption & 12.9 & 4.6 & 16.4 & 5.4 \\
Description & 13.1 & 4.5 & 16.6 & 5.5 \\
CoE Text$^*$ & 13.3 & 4.6 & 16.9 & 5.6 \\
\toprule

\multicolumn{5}{l}{\textbf{(e) Impact of Semantic-Aware Text-Visual Alignment}}\\

\multicolumn{5}{l}{\textit{Alignment Similarity Metric}} \\
Euclidean & 11.5 & 4.3 & 14.5 & 5.3 \\
Manhattan & 13.5 & 5.3 & 17.7 & 6.2 \\
\cdashline{1-5} 
\multicolumn{5}{l}{\textit{Alignment Strategy}} \\
VV (Video-Video Only) & 11.3 & 3.5 & 13.8 & 4.2 \\
VT (Video-Text Only) & 12.4 & 4.9 & 13.4 & 5.3 \\
VV+VT (Summation) & 12.7 & 4.0 & 15.8 & 4.8 \\
\toprule

\multicolumn{5}{l}{\textbf{(f) Impact of CoE Text Quality and Encoding}} \\
\multicolumn{5}{l}{\textit{Text Quality}} \\
Key-content corruption (low-quality CoE) & 13.8 & 5.0 & 17.4 & 6.9 \\
Causal-order reversal (erroneous CoE) & 12.9 & 4.4 & 13.6 & 4.2 \\
\cdashline{1-5}
\multicolumn{5}{l}{\textit{Text Encoding Strategy}} \\
Concatenated single text & 13.7 & 4.8 & 16.9 & 6.6 \\
\toprule

\multicolumn{5}{l}{\textbf{(g) Impact of Background Snippet Mask}} \\
\multicolumn{5}{l}{\textit{Video-Video Alignment}} \\
$\mathcal{M}^v$ only & 10.9 & 3.4 & 13.1 & 3.6 \\
$\mathcal{M}^v\odot\mathcal{M}^m$ & 11.3 & 3.5 & 13.8 & 4.2 \\
\cdashline{1-5} 
\multicolumn{5}{l}{\textit{Video-Text Alignment}} \\
$\mathcal{M}^{vt}$ only & 11.7 & 3.8 & 12.4 & 4.1 \\
$\mathcal{M}^{vt}\odot\mathcal{M}^m$ & 12.4 & 4.9 & 13.4 & 5.3 \\
\cdashline{1-5} 
\multicolumn{5}{l}{\textit{Combined Alignment}} \\
$\mathcal{M}^v\odot\mathcal{M}^{vt}$ (w/o $\mathcal{M}^m$) & 13.6 & 5.2 & 16.5 & 6.2 \\
\toprule

Ours (on THUMOS14) & \textbf{14.1} & \textbf{5.4} & \textbf{18.2} & \textbf{7.3} \\
\bottomrule
\end{tabular}
\end{table}

\textbf{Impact of Semantic-Temporal Pyramid Encoder.} 
We evaluate the impact of different variants of STPE in Table~\ref{tab:combined_results} (c). The following variants are considered:
1) \textit{w/o STPE}: remove the STPE and use only the backbone;
2) \textit{w/o TP}: utilize only the semantic pyramid block;
3) \textit{w/o SP}: utilize only the temporal pyramid block; 
4) \textit{Transformer}: replace STPE with Transformer~\cite{vaswani2017attention}, which has comparable parameters for feature extraction.
The results indicate that removing the semantic or temporal pyramid block leads to a decrease in performance. Notably, completely removing STPE causes a significant drop. In comparison, our approach outperforms Transformer, demonstrating that robust feature extraction from both semantic and temporal dimensions significantly enhances localization performance. To further separate the contribution of the textual branch from the rest of the architecture, we also report a text-free variant, Ours$^*$, which removes the textual branch and retains only the visual branch, as shown in Table~\ref{tab:main_table}. Incorporating the textual branch consistently improves performance over Ours$^*$ across most settings on both datasets, confirming the effectiveness of the textual branch for localization.

\textbf{Impact of Different Textual Content.} We compare our text generation approach with alternative methods in Table~\ref{tab:combined_results} (d). The following text categories are evaluated: 1) \textit{Prompt}: category-based prompts, \emph{e.g.}, ``A video of class basketball dunk''; 2) \textit{Caption}: frame-level captions generated by CoCa~\cite{yu2205coca}; 3) \textit{Video description}: detailed descriptions produced by VideoChat~\cite{li2024videochat}; 4) \textit{CoE Text$^*$}: the CoE text solely produced by VideoChat.
Among the evaluated methods, \textit{prompt} lacks specificity regarding video content, while \textit{caption}  fails to capture long-term temporal dependencies within the action sequence. Although \textit{video description} provides temporal information, it often introduces noise through redundant environment details. Since VideoChat is not a reasoning model, \textit{CoE Text$^*$} cannot effectively uncover logical relationships, resulting in lower quality of the generated CoE text. In contrast, our proposed CoE Text generated by VLM and LLM effectively extracts relevant context and exploits both temporal and causal relationships, thereby enhancing cross-modal alignment and localization accuracy. As demonstrated in Table~\ref{tab:combined_results} (d), our method consistently outperforms other approaches across multiple evaluation metrics.
Furthermore, we analyze whether CoE quality affects localization by perturbing the original CoE text, as shown in Table~\ref{tab:combined_results} (f). Key-content corruption weakens action-relevant evidence, while causal-order reversal introduces erroneous causal relations. Both perturbations reduce performance, confirming that accurate and causally ordered CoE text provides useful semantic guidance. We also compare our sentence-wise CoE encoding against concatenating all sentences into a single passage encoded as one vector, as shown in Table~\ref{tab:combined_results} (f). The concatenated single-text variant performs worse in both 1-shot and 5-shot settings, indicating that single-text encoding hurts sentence-level temporal correspondence and weakens fine-grained semantics, which supports the sentence-wise encoding used in our proposed method.

\textbf{Impact of Semantic-Aware Text-Visual Alignment.} 
We conduct an ablation study on the choice of similarity metric and alignment strategies, with results reported in Table~\ref{tab:combined_results} (e).
First, we evaluate different metrics for computing the alignment maps between query and support features. We compare cosine similarity against Euclidean and Manhattan distances. The experimental results clearly show that cosine similarity significantly outperforms the other two, showing it focuses on the orientation of the feature vectors (\emph{i.e.}, semantic content) rather than their magnitude, making it more robust for capturing semantic relationships between features. Subsequently, we evaluate the impact of three different alignment strategies.
1) \textit{VV}: align only the query video features with the support video features by operation $\mathcal{S}$;
2) \textit{VT}: align the query video features with support features that have been aligned with text by operation $\mathcal{S}$;
3) \textit{VV+VT}: combining both strategies (\textit{VV} and \textit{VT}) by summation. 
The results indicate that incorporating textual semantic information for alignment outperforms methods relying solely on video information. Among the strategies evaluated, our approach which leverages both video and text information yields the best results. Furthermore, using multiplication in our method is more effective than direct addition, as it refines the incorrect alignment in $\mathcal{M}^v$ by multiplying it with the correct alignment score in $\mathcal{M}^{vt}$ derived from the auxiliary textual information. In contrast, direct addition inherently accumulates misalignment errors, which degrade accuracy at higher IoU thresholds and consequently reduce the mAP.
We further evaluate the background mask $\mathcal{M}^m$ under different alignment-source configurations in Table~\ref{tab:combined_results} (g). Adding $\mathcal{M}^m$ consistently improves performance for video-video alignment, video-text alignment, and their combination. This confirms that suppressing support background snippets is complementary to both alignment sources.

\begin{figure}
    \centering
    \includegraphics[width=0.8\linewidth]{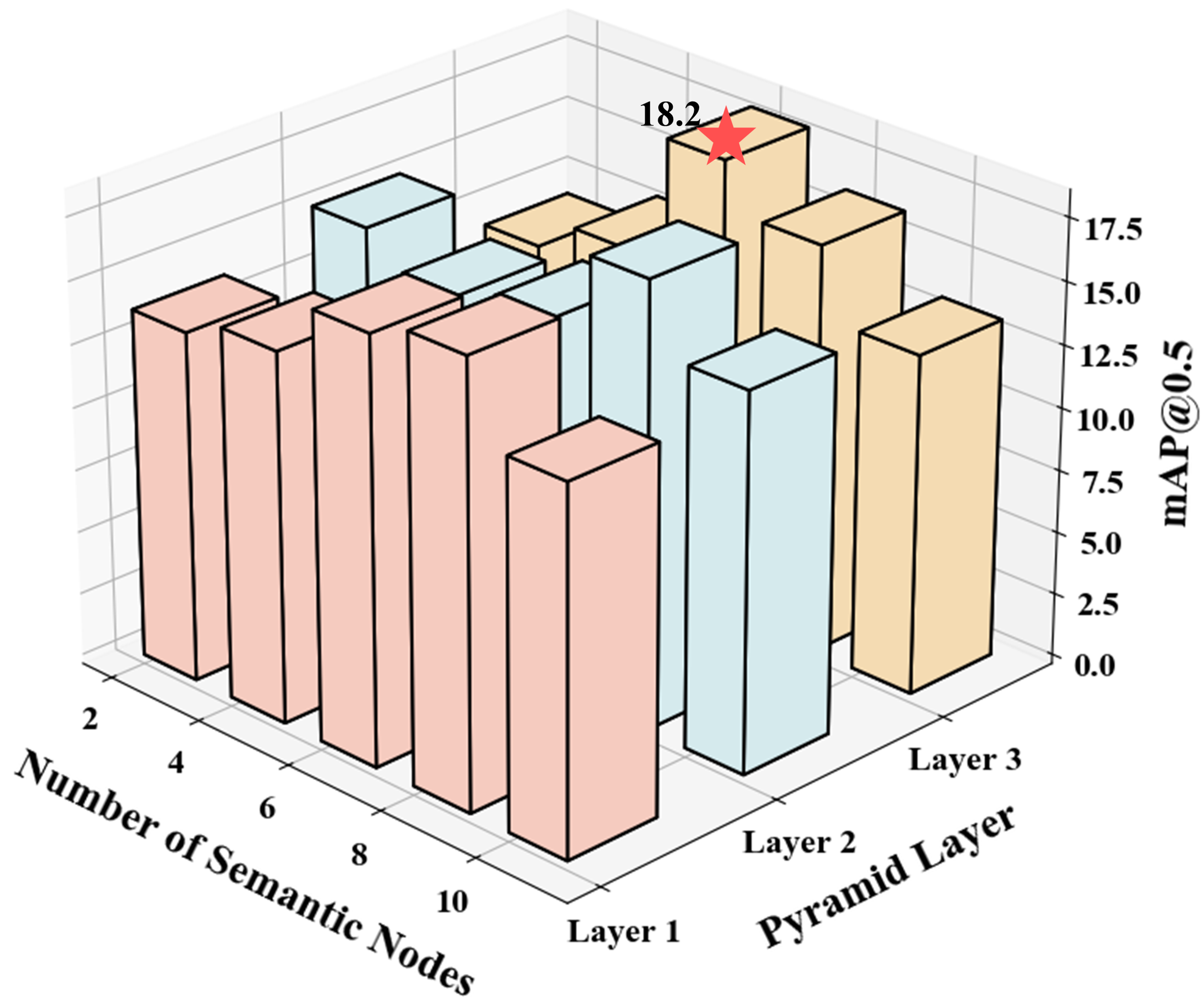}
    \caption{Ablation study on hyperparameters of Semantic Pyramid: number of layers and semantic nodes. Best performance is marked with a red star.}
    \vspace{-4mm}
    \label{fig:semantic}
\end{figure}

\textbf{Different Semantic Pyramid Layers and Semantic Nodes.}
We evaluate the impact of the number of semantic pyramid layers and nodes (snippets) performing semantic attention operations at each layer within the Semantic-Temporal Pyramid Encoder (STPE). The results of mAP@0.5 under the multi-instance scenario of THUMOS14 are shown in Figure~\ref{fig:semantic}. We fix the number of STPE blocks to 2 and vary the number of pyramid layers for semantic attention operations to range from 1 to 3. When the number of layers is fixed, adding more semantic nodes leads to a gradual peak on mAP@0.5 before it starts to decline. This decline may result from the heightened likelihood of both foreground and background snippets simultaneously engaging in semantic attention operations, which diminishes the model's ability to capture variance within the action and fails to distinguish between foreground and background snippets. Our proposed method achieves optimal performance when the number of semantic nodes is 6 and the number of layers is 3, as indicated by the red mark in Figure~\ref{fig:semantic}.

\begin{table}[t]
\centering
\caption{
    The impact of different backbones on the ActivityNet1.3 dataset.
}
\setlength{\tabcolsep}{3.5mm} 
\label{tab:ablation_backbone}
\renewcommand{\arraystretch}{1.1}
\begin{tabular}{lcccc}
\toprule
\multirow{2}{*}{Method} & \multicolumn{2}{c}{Single-instance} & \multicolumn{2}{c}{Multi-instance} \\
\cmidrule(lr){2-3} \cmidrule(lr){4-5}
& 1-shot & 5-shot & 1-shot & 5-shot \\
\midrule
Lee et al.~\cite{lee2023few} & 63.1 & 67.5 & 49.4 & 54.6 \\
Zhang et al.~\cite{CGCN2025} & 66.6 & 67.2 & - & - \\
Ours (ViViT) & 66.7 & 67.7 & 52.4 & 53.3 \\
Ours (C3D+STPE) & \underline{69.6} & \underline{71.5} & \underline{54.9} & \underline{58.7} \\
Ours (ViViT+STPE) & \textbf{71.3} & \textbf{75.1} & \textbf{58.6} & \textbf{60.1} \\
\bottomrule
\end{tabular}
\end{table}
% \vspace{-3mm}

\textbf{Impact of Different Backbones.} 
To ensure a fair comparison with prior works, we adopt the conventional C3D~\cite{7410867} as the backbone for our main experiments. 
Moreover, we also conduct an additional evaluation on the ActivityNet1.3 dataset using a more advanced Transformer-based backbone, ViViT~\cite{vivit}, with the mAP@0.50 results presented in Table~\ref{tab:ablation_backbone}.
As the table shows, merely replacing the backbone with ViViT already establishes a strong baseline. 
Crucially, when our STPE module is further applied on the features extracted by ViViT, the performance is significantly enhanced across all settings. The results not only validate the direct benefits of using a superior backbone, but also indirectly demonstrate the effectiveness of our proposed STPE module.

\textbf{Impact of CLIP.} 

\begin{figure}
    \centering
    \includegraphics[width=1.0\linewidth]{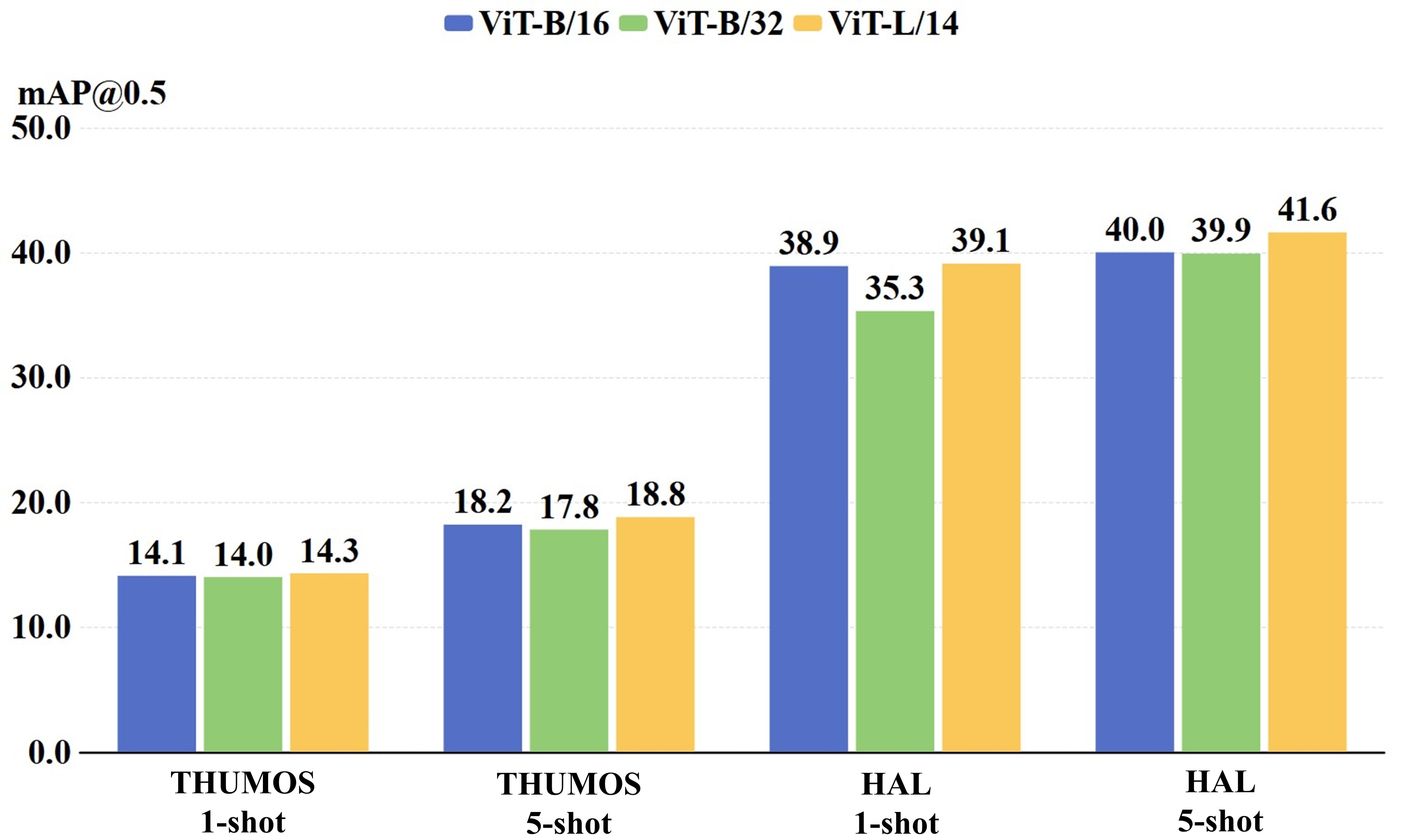}
    \vspace{-5mm}
    \caption{The impact of different CLIP variants compared across the THUMOS14 and HAL datasets.}
    \label{fig:clip}
\end{figure}
We evaluate the impact of different CLIP models on the localization performance. With the VLM and LLM fixed as VideoChat and DeepSeek-R1, we employ three CLIP models: clip-vit-base-patch16, clip-vit-base-patch32, and clip-vit-large-patch14 to perform textual encoding of both frame-level captions and CoE texts. The results of mAP@0.5 under the multi-instance scenario of THUMOS14 and HAL datasets are shown in Figure \ref{fig:clip}. The clip-vit-base-patch16 captures finer visual details and achieves more granular alignment in CLIP’s semantic space than clip-vit-base-patch32. Its text encoder better captures detailed textual descriptions and maps them to the semantic alignment space, assisting the visual model in detecting subtle actions and thereby improving performance. Although the larger CLIP model achieves slightly better performance on certain metrics, its overall effectiveness is comparable to or even inferior to that of clip-vit-base-patch16, especially when considering the higher resource consumption. Therefore, we adopt clip-vit-base-patch16 as the default CLIP model in subsequent experiments.

\begin{table}[t]
\centering
\caption{
    The impact of different VLMs and LLMs on the THUMOS14 and HAL datasets. ``*'' denotes reasoning model and ``Q'' represents ``Qwen''.
}
\label{table:vlm_llm_combined}
\setlength{\tabcolsep}{0.1mm} 
\renewcommand{\arraystretch}{1.05} 

\begin{tabular}{llcccccccc}
\toprule
\multirow{2}{*}{VLM} & \multirow{2}{*}{LLM} & \multicolumn{4}{c}{THUMOS14} & \multicolumn{4}{c}{HAL} \\
\cmidrule(lr){3-6} \cmidrule(lr){7-10}
& & \multicolumn{2}{c}{1-shot} & \multicolumn{2}{c}{5-shot} & \multicolumn{2}{c}{1-shot} & \multicolumn{2}{c}{5-shot} \\
\cmidrule(lr){3-4} \cmidrule(lr){5-6} \cmidrule(lr){7-8} \cmidrule(lr){9-10}
& & 0.5 & Mean & 0.5 & Mean & 0.5 & Mean & 0.5 & Mean \\
\midrule
\multirow{2}{*}{\shortstack{Q2.5-VL\\7B}}
& Q3-30B(no think) & 13.4          & 4.2          & 14.1          & 4.0          & 38.0          & 23.6         & 39.3          & 24.3         \\
& Q3-30B(think)$^*$ & \textbf{14.9} & 4.3          & 18.7          & 5.8          & 39.5          & 23.8         & 41.3          & 24.8         \\
\midrule
\multirow{4}{*}{\shortstack{VideoChat\\7B}} 
& Q3-30B(no think) & 14.3          & 4.0          & 15.0          & 4.6          & 38.1          & 25.1         & 38.9          & 24.0         \\
& Q3-30B(think)$^*$ & 14.6          & 4.3          & \textbf{18.8} & 5.9          & \textbf{39.9} & 24.1         & \textbf{42.6} & \textbf{27.0} \\
& Q2.5-Max        & 13.0          & 4.0          & 15.1          & 4.4          & 37.8          & 24.6         & 39.9          & 25.4         \\
& DeepSeek-R1-70B$^*$    & 14.1          & \textbf{5.4} & 18.2          & \textbf{7.3} & 38.9          & \textbf{25.2} & 40.0          & 26.7         \\
\bottomrule
\end{tabular}
\end{table}

\textbf{Impact of VLMs and LLMs.} With CLIP fixed as clip-vit-base-patch16, we conduct a comparative analysis of the impact of different VLMs and LLMs on overall performance. For the VLM, we select Qwen2.5-VL-7B-Instruct~\cite{qwen25vl} and VideoChat-Flash-Qwen2-7B-res448~\cite{li2024videochat}; for the LLM, we choose Qwen3-30B-A3B~\cite{yang2025qwen3}, DeepSeek-R1-Distill-Llama-70B~\cite{guo2025deepseek} and Qwen2.5-Max~\cite{qwen25}. Among these LLMs, Qwen3-30B-A3B supports dynamic switching between standard and reasoning modes during inference. When operating in reasoning mode, it is considered a reasoning model. All reasoning models are marked with a $^*$. We conduct experiments on the THUMOS14 and HAL datasets under the multi-instance scenario, and report the mAP@0.5 and mean mAP results for both 1-shot and 5-shot settings in Table~\ref{table:vlm_llm_combined}. As shown in the table, reasoning models consistently outperform non-reasoning models, and the reasoning capability of the LLM has a more significant impact on performance than the choice of VLM. For instance, Qwen3-30B-A3B achieves better results in reasoning (think) mode than in standard (no think) mode, demonstrating the positive effect of reasoning-based text generation. Moreover, despite having fewer parameters, DeepSeek-R1-Distill-Llama-70B outperforms Qwen-Max, further validating that reasoning-generated text can more effectively enhance model performance.

\begin{figure*}[t]
    \centering
    \subfloat[]{
        \includegraphics[width=0.9\linewidth]{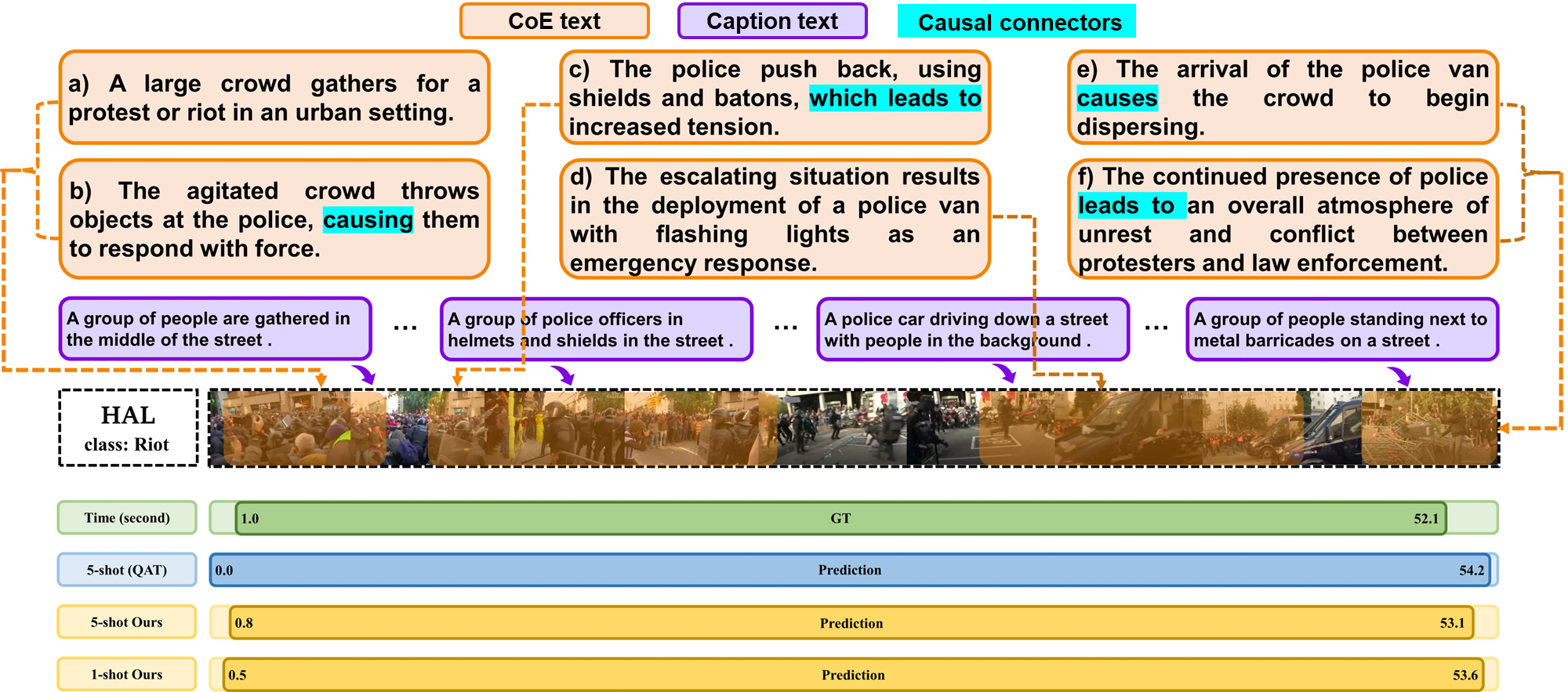}
        \label{fig:qualitative_result_1}
    }
    \\ 
    \vspace{-3mm} 
    \subfloat[]{
        \includegraphics[width=0.87\linewidth]{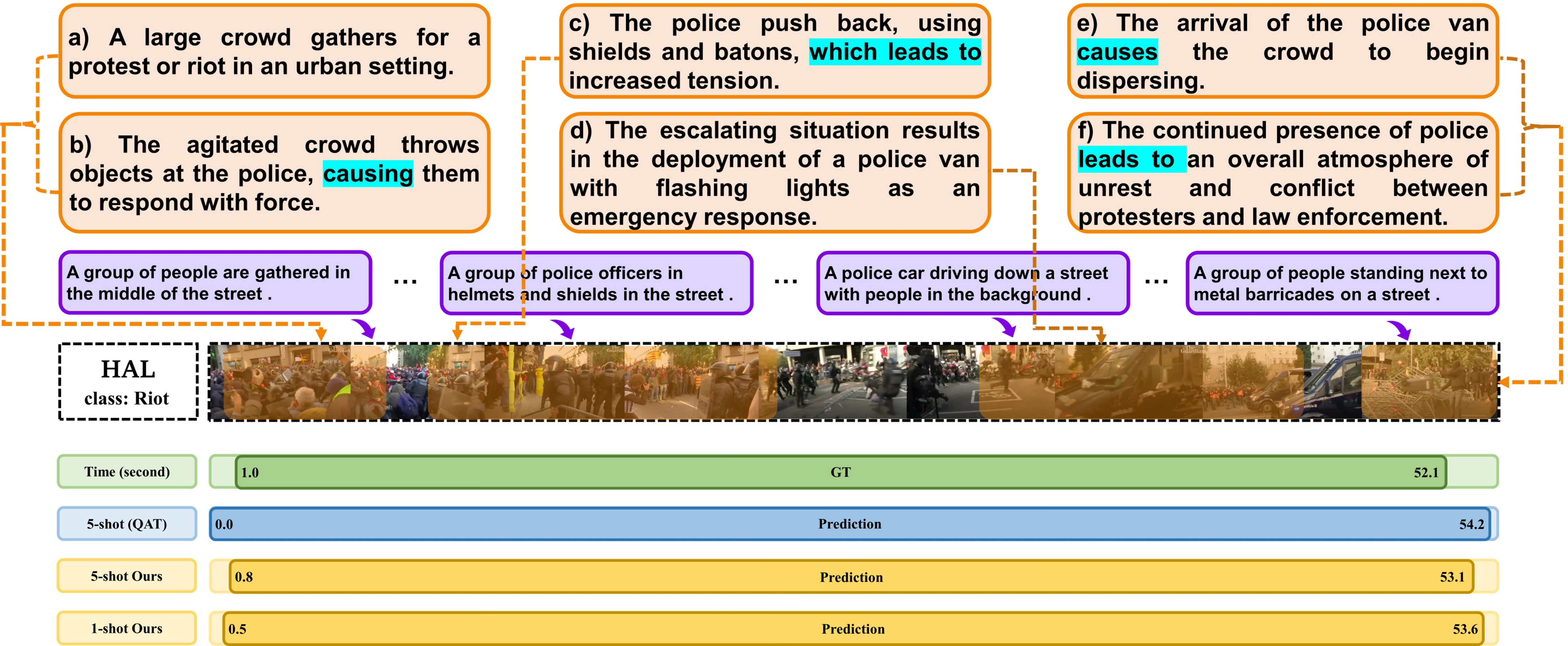}
        \label{fig:qualitative_result_2}
    }
    \caption{Qualitative comparisons of our method against QAT~\cite{nag2021few} and Ground Truth. Purple boxes denote standard captions, and orange boxes represent our generated CoE text.(a) Visual results on ``Hurling'' from ActivityNet 1.3. While standard captions provide only brief descriptions, the CoE text offers a comprehensive narrative that better defines the action context. (b) Visual results on ``Riot'' from the HAL dataset. Unlike captions that only offer static descriptions, the CoE text captures the more meaningful causal relation (highlighted in \textcolor{cyan}{cyan}, \emph{e.g.}, ``causing'', ``leads to'') of the anomaly.}
    \label{fig:qualitative_comparisons}
    \vspace{-3mm}
\end{figure*}

\subsection{Qualitative analysis.} 
To prove the effectiveness of our method, we show qualitative results of one case from the ActivityNet1.3 in Figure~\ref{fig:qualitative_result_1} and one case from the HAL dataset in Figure~\ref{fig:qualitative_result_2}. We observe that our method locates the action snippets more accurately than QAT~\cite{nag2021few} under a 5-shot setting and maintains comparable performance under the 1-shot setting. 
This improvement is largely attributed to our CoE reasoning. Unlike the standard caption that describes an isolated state, our CoE text explicitly expresses the causal relation and the evidence of the anomaly by logical connectors (like ``causing'' and ``leads to'' as Figure~\ref{fig:qualitative_result_2} shows). This structured CoE text serves as a semantic guide, enabling the model to logically connect sequential sub-events ($e_i \rightarrow e_{i+1}$) and identify the critical start and end of the anomaly, thus resulting in more precise temporal boundaries.

\section{Conclusion}
In this paper, we presented a novel few-shot TAL method to enhance localization performance by integrating textual semantic information. 
First, we designed a CoE reasoning method to generate textual descriptions that can express temporal dependencies and causal relationships between actions. 
Then, a novel few-shot learning framework was designed to capture hierarchical action commonalities and variations by aligning query and support videos. Extensive experiments demonstrate the effectiveness and superiority of our proposed method. In the future, we will leverage our proposed method into more vertical fields, such as social security governance.

\bibliographystyle{IEEEtran}
\bibliography{main}

\vspace{-15mm}
\begin{IEEEbiography}[{\includegraphics[width=1in,height=1in,keepaspectratio]{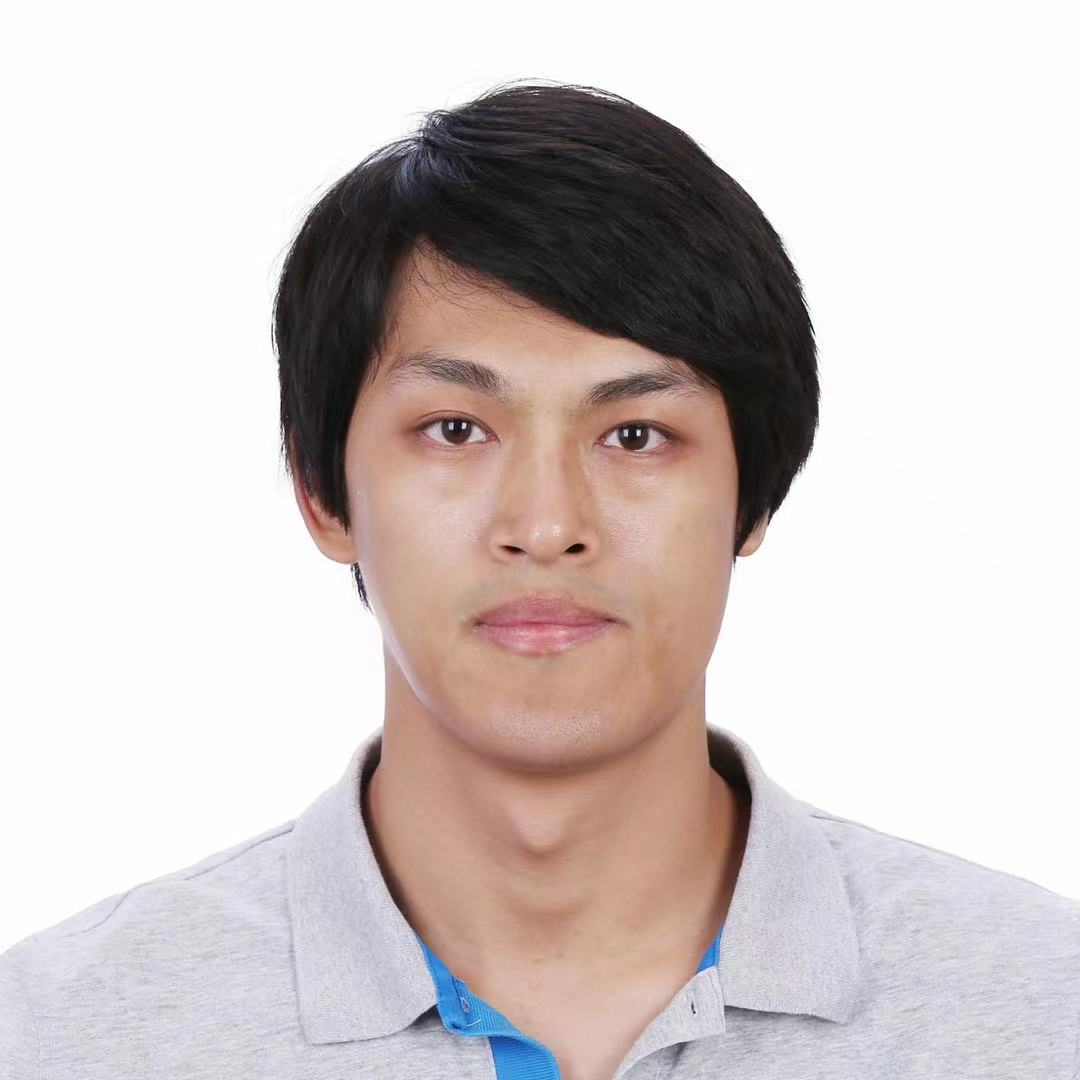}}] {Mengshi Qi} (Member, IEEE) is currently a Professor with the Beijing University of Posts and Telecommunications, Beijing, China. He received Ph.D. degrees in computer science from Beihang University, Beijing, China, in 2019. His research interests include machine learning and computer vision, especially scene understanding, 3D reconstruction, and multimedia analysis. He has published more than 50 papers in top journals (such as IEEE TIP, TPAMI, TMM, TCSVT, TIFS) and top conferences (such as IEEE CVPR, ICCV, ECCV, ACM Multimedia, AAAI, NeurIPS). He also has served as Area Chair of ICME 2024-2025, Senior PC Member of AAAI 2023-2025 and IJCAI 2021/2023-2025, and the Guest Editor for IEEE Transactions on Multimedia.
\end{IEEEbiography}
\vspace{-15mm}

\begin{IEEEbiography}[{\includegraphics[width=1in,height=1in,keepaspectratio]{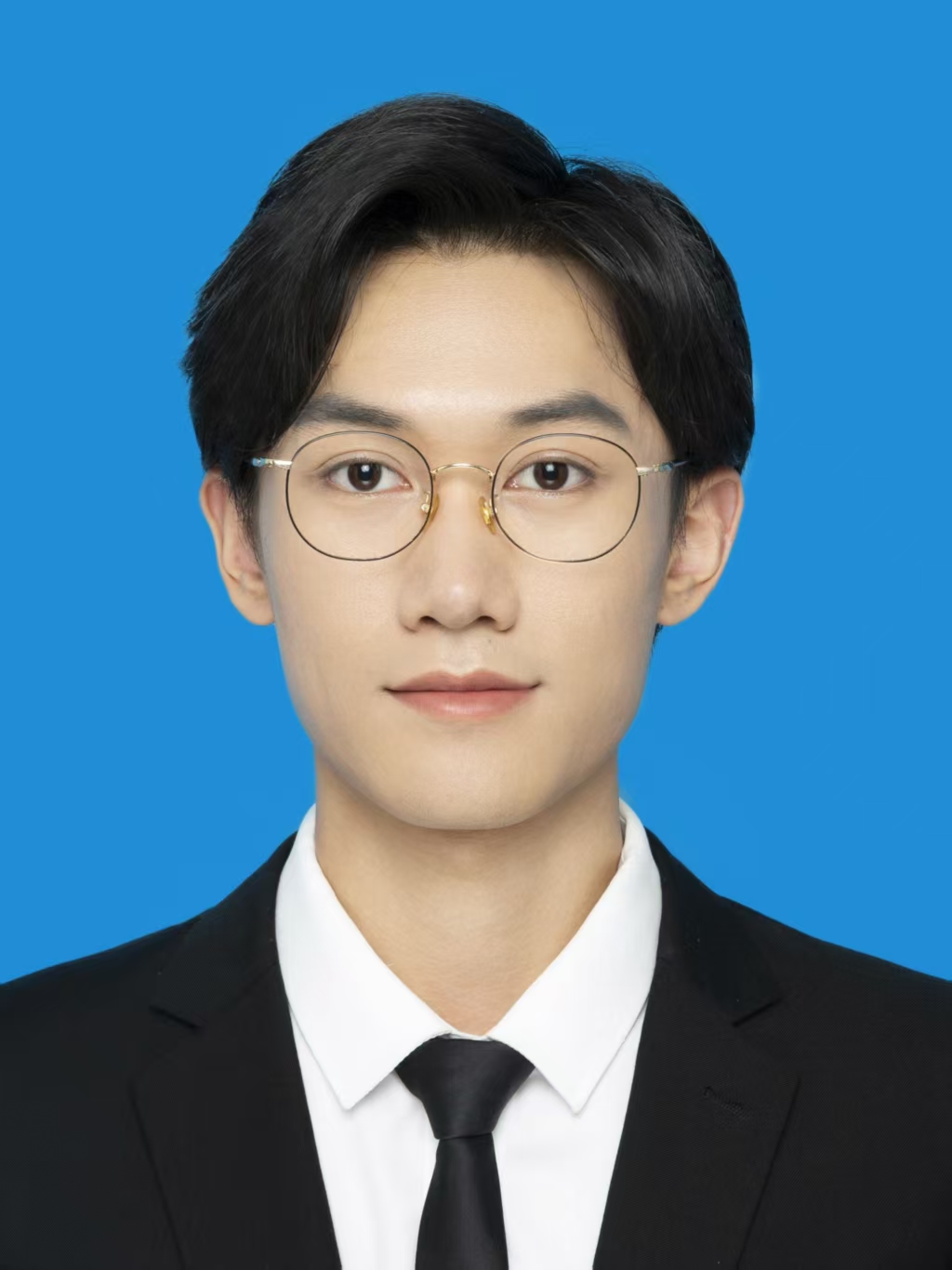}}] {Hongwei Ji} is currently working toward the master’s degree at the Beijing University of Posts and Telecommunications, Beijing, China. His research interests include multimodal learning and computer vision, particularly temporal action localization.
\end{IEEEbiography}
\vspace{-15mm}

\begin{IEEEbiography}[{\includegraphics[width=1in,height=1in,keepaspectratio]{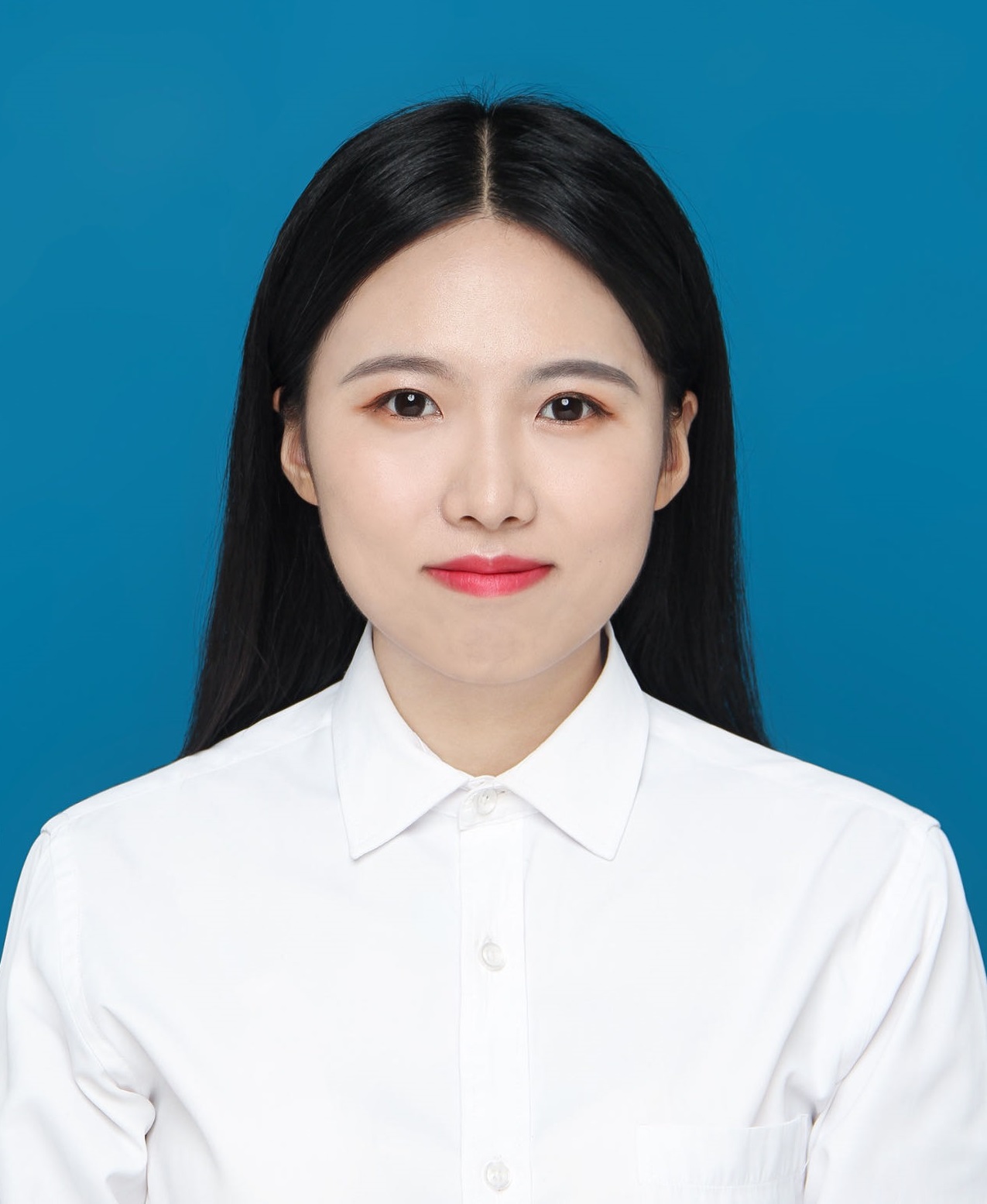}}] {Wulian Yun} is currently pursuing a Ph.D. degree at Beijing University of Posts and Telecommunications, China. Her research interests include machine learning and computer vision, especially temporal action localization and video understanding. 
\end{IEEEbiography}
\vspace{-15mm}

\begin{IEEEbiography}[{\includegraphics[width=1in,height=1in,keepaspectratio]{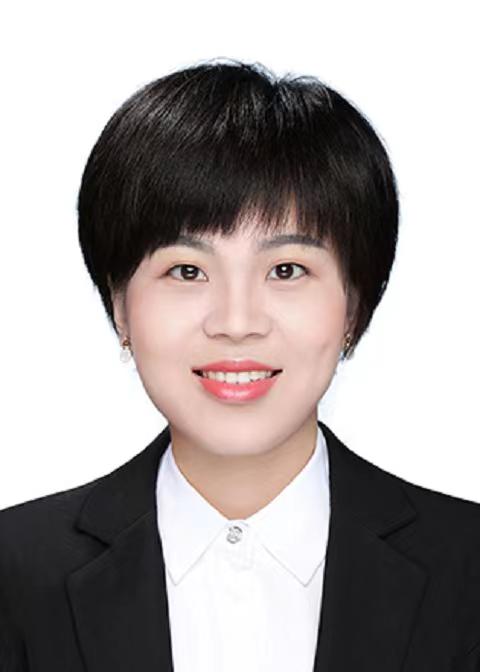}}] {Xianlin Zhang} is an Assistant Professor at Beijing University of Posts and Telecommunications, China. Her current research focuses on generative artificial intelligence and video analysis and understanding. She received her Ph.D. degree from Beijing University of Posts and Telecommunications in 2019. To date, she has authored or co-authored more than 20 publications in prestigious journals and conferences, including CVPR, TMM, PR, ACM TOMM.
\end{IEEEbiography}
\vspace{-15mm}

\begin{IEEEbiography}[{\includegraphics[width=1in,height=1in,keepaspectratio]{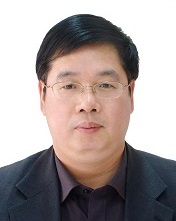}}]{Huadong Ma} (Fellow, IEEE) received Ph.D. degree in computer science from the Institute of Computing Technology, Chinese Academy of Science, Beijing, China, in 1995. He is currently a Professor of School of Computer Science, Beijing University of Posts and Telecommunications, Beijing, China. His current research interests include Internet of Things, sensor networks, and multimedia computing. He has authored more than 300 papers in journals, such as ACM/IEEE Transactions or conferences, such as ACM MobiCom, ACM SIGCOMM, IEEE INFOCOM, and five books. He was the recipient of the Natural Science Award of the Ministry of Education, China, in 2017, 2019 Prize Paper Award of IEEE TRANSACTIONS ON MULTIMEDIA, 2018 Best Paper Award from IEEE MULTIMEDIA, Best Paper Award in IEEE ICPADS 2010, Best Student Paper Award in IEEE ICME 2016 for his coauthored papers, and National Funds for Distinguished Young Scientists in 2009. He was/is an Editorial Board Member of the IEEE TRANSACTIONS ON MULTIMEDIA, IEEE INTERNET OF THINGS JOURNAL, ACM Transactions on Internet of Things.
\end{IEEEbiography}

\end{document}

\title{Supplementary for Chain-of-Evidence Multimodal Reasoning \\for Few-shot Temporal Action Localization}

\author{Mengshi Qi,~\IEEEmembership{Member,~IEEE,}
        Hongwei Ji,
        Wulian Yun,
        Xianlin Zhang,
        Huadong Ma,~\IEEEmembership{Fellow,~IEEE} 

\thanks{M. Qi, H. Ji, W. Yun, X. Zhang and H. Ma are with State Key Laboratory of Networking and Switching Technology, Beijing University of Posts and Telecommunications, China.}
}

\IEEEpubid{}

\maketitle
\section{Training and inference processes }
\noindent
For a better understanding of our few-shot temporal action localization task, we present the training and testing processes below. 
For the training process, we first randomly select a class $C$ from the training subset $D_{train}$. Subsequently, we select $1+K$ samples as query and support from $C$. For the support video, we generate caption text descriptions on each frame and CoE text descriptions via VLM and LLM. Then, we perform feature extraction on both the videos and the text. Specifically, for both the query video and each support video, we extract snippet features $\mathcal{V}^q$ and $\mathcal{V}^s$ using a fixed pre-trained backbone, and the corresponding support text is processed using the CLIP Text Encoder to extract features $\mathcal{F}^{cap}$ and $\mathcal{F}^{CoE}$. 
These extracted video features are then passed through the STPE to further capture features $\mathcal{F}^q$ and $\mathcal{F}^s$,
text features $\mathcal{F}^{cap}$ and $\mathcal{F}^{CoE}$ are aligned through a cross-attention operation to generate the final text feature of the support set. Next, we align the support video features $\mathcal{F}^s$ with the textual features $\mathcal{F}^s$ by a video-text alignment block.  
Then, the query features $\mathcal{F}^t$, support features $\mathcal{F}^s$. The video-text aligned features $\mathcal{F}^{\hat{s}}$ are fed into a semantic-aware text-visual alignment module with Eq.(4) $\sim$ Eq.(7) in our main paper. The aligned features are subsequently passed to the prediction head to perform action localization.
Finally, we compute the loss through Eq.(8) $\sim$ Eq.(10) in our main paper to optimize our model and update all parameters.

For the inference process, it is similar to the training process. The difference is that after obtaining the predictions, we need to select continuous snippets as proposals based on different thresholds in $\mathcal{T}$ and then apply soft-NMS on all proposals.  
The training and testing processes are illustrated in the algorithms below.

\begin{algorithm}[hbt!]
\small
\caption{The Training Process of Our Method}
\label{alg:training}
\begin{algorithmic}[1] 
    \Require Training Dataset $D_{\text{train}}$, Epochs $E$, Episodes per Epoch $N_{ep}$, Shots $K$, Model ${M}_\theta$
    \Ensure Optimized Model ${M}_\theta$
    
    \For{$epoch = 1 \to E$}
        \For{$episode = 1 \to N_{ep}$}
            \State \Comment{Sample a task (episode)}
            \State Sample a class $C$ from $D_{\text{train}}$
            \State Sample query video $\mathcal{V}^q$ and support set $\{\mathcal{V}_i^s\}_{i=1}^K$ from $C$
            
            \State \Comment{Text Feature Extraction (offline)}
            \State Generate captions and CoE text for $\{\mathcal{V}_i^s\}$
            \State $\mathcal{F}^{\text{cap}}, \mathcal{F}^{\text{CoE}} \gets \text{CLIP-Text-Encoder}(\text{texts})$
            
            \State \Comment{Forward Pass}
            \State $\mathcal{F}^q, \mathcal{F}^s \gets \text{BackboneAndSTPE}(\mathcal{V}^q, \{\mathcal{V}_i^s\})$
            \State $\mathcal{F}^t \gets \text{CrossAttention}(\mathcal{F}^{\text{cap}}, \mathcal{F}^{\text{CoE}})$
            \State $\mathcal{M}^v \gets \text{CosineSimilarity}(\mathcal{F}^q, \mathcal{F}^s)$ \Comment{Eq. (4)}
            \State $\mathcal{F}^{\hat{s}} \gets \text{Fuse}(\mathcal{F}^s, \mathcal{F}^t)$ \Comment{Eq. (5)}
            \State $\mathcal{M}^{vt} \gets \text{CosineSimilarity}(\mathcal{F}^q, \mathcal{F}^{\hat{s}})$ \Comment{Eq. (6)}
            \State $\mathcal{M} \gets \mathcal{M}^v \odot \mathcal{M}^{vt} \odot \mathcal{M}^{m}$ \Comment{Eq. (7)}
            \State $\hat{p} \gets \text{PredictionHead}(\mathcal{M})$
            
            \State \Comment{Compute Loss and Update Model}
            \State $\mathcal{L} \gets \text{ComputeLoss}(\hat{p}, y^q)$ \Comment{Using Eq. (8-10)}
            \State Update model parameters $\theta$ by backpropagating $\mathcal{L}$
        \EndFor
    \EndFor
\end{algorithmic}
\end{algorithm}

\begin{algorithm}[hbt!]
\small
\caption{The Inference Process of Our Method}
\label{alg:inference}
\begin{algorithmic}[1]
    \Require Test Query Video $\mathcal{V}^q$, Support Set $\{\mathcal{V}_i^s, y_i^s\}_{i=1}^K$, Frozen Model ${M}_\theta$, Confidence Threshold $\tau_{\text{conf}}$, NMS Threshold $\tau_{\text{nms}}$
    \Ensure Final Action Proposals $P_{\text{final}}$
    
    \State \Comment{Feature Extraction}
    \State Generate/Load captions and CoE text for $\{\mathcal{V}_i^s\}$
    \State $\mathcal{F}^{\text{cap}}, \mathcal{F}^{\text{CoE}} \gets \text{CLIP-Text-Encoder}(\text{texts})$
    \State $\mathcal{F}^q, \mathcal{F}^s \gets \text{BackboneAndSTPE}(\mathcal{V}^q, \{\mathcal{V}_i^s\})$
    \State $\mathcal{F}^t \gets \text{CrossAttention}(\mathcal{F}^{\text{cap}}, \mathcal{F}^{\text{CoE}})$

    \State \Comment{Compute Snippet-level Probabilities}
    \State $\mathcal{M}^v \gets \text{CosineSimilarity}(\mathcal{F}^q, \mathcal{F}^s)$
    \State $\mathcal{F}^{\hat{s}} \gets \text{Fuse}(\mathcal{F}^s, \mathcal{F}^t)$
    \State $\mathcal{M}^{vt} \gets \text{CosineSimilarity}(\mathcal{F}^q, \mathcal{F}^{\hat{s}})$
    \State $\mathcal{M}^{m} \gets \text{GenerateBackgroundMask}(\{y_i^s\})$
    \State $\mathcal{M} \gets \mathcal{M}^v \odot \mathcal{M}^{vt} \odot \mathcal{M}^{m}$
    \State $\hat{p} \gets \text{PredictionHead}(\mathcal{M})$ \Comment{Snippet-level probabilities}
    
    \State \Comment{Generate Proposals}
    \State $P_{\text{raw}} \gets \emptyset$ \Comment{Initialize raw proposal set}
    \State Identify consecutive snippets in $\mathcal{V}^q$ where probabilities $\hat{p} > \tau_{\text{conf}}$
    \State Add identified snippets as proposals to $P_{\text{raw}}$
    \State Filter out proposals that are too short
    
    \State \Comment{Refine Proposals}
    \State $P_{\text{final}} \gets \text{Soft-NMS}(P_{\text{raw}}, \tau_{\text{nms}})$
    \State \Return $P_{\text{final}}$
\end{algorithmic}
\end{algorithm}

\begin{algorithm}[hbt!]
\small
\caption{Pipeline for Generating and Verifying CoE Text}
\label{alg:cot_pipeline_separated}
\begin{algorithmic}[1]
    \Require Video $\mathcal{V}$, VLM ${M}_{vlm}$, LLM ${M}_{llm}$, CLIP Textual Encoder $M_{ct}$, CLIP Visual Encoder $M_{cv}$, Prompts Library $\mathcal{P}_{1,2,3}$, Max Retries $N_{retry}$, Threshold $\alpha$
    \Ensure Verified CoE text $o_3$
    
    \State \Comment{--- Stage 1: Generate and Verify Detailed Description ---}
    \State $p_1 \gets \text{SelectPrompt}(\mathcal{P}_1)$, $x_0 \gets \mathcal{V}$
    \State $o_1 \gets \text{GenerateAndVerify}(p_1, \mathcal{M}_{vlm}, \alpha, N_{retry}, \text{context}=x_0)$
    \If{$o_1$ is marked for human review} \Return Failure \EndIf
        
    \State \Comment{--- Stage 2: Extract and Verify Key Events ---}
    \State $p_2 \gets \text{SelectPrompt}(\mathcal{P}_2)$, $x_1 \gets o_1$
    \State $o_2 \gets \text{GenerateAndVerify}(p_2, \mathcal{M}_{vlm}, \alpha, N_{retry}, \text{context}=x_1)$
    \If{$o_2$ is marked for human review} \Return Failure \EndIf
    
    \State \Comment{--- Stage 3: Generate and Verify CoE Text ---}
    \State $p_3 \gets \text{SelectPrompt}(\mathcal{P}_3)$, $x_2 \gets \{o_1,o_2\}$
    \State $o_3 \gets \text{GenerateAndVerify}(p_3, \mathcal{M}_{llm}, \alpha, N_{retry}, \text{context}=x_2)$
    \State \Return $o_3$

    \Statex
    \Procedure{AutoFilter}{Text $o$}
        \If{$o$ \text{is too short or repeat}}
            \State \Return False
        \EndIf
        \State \Return True
    \EndProcedure
    
    \Statex
    \Procedure{Verify}{Video $\mathcal{V}$, Text $o$, Threshold $\alpha$}
        \If{$\text{not AutoFilter}(o)$}
            \State \Return (False, \{\text{too short or repeat}\})
        \EndIf
        \State $C \gets \text{ParseText}(o)$; $F \gets \text{SampleFrames}(\mathcal{V})$
        \State $S \gets \text{CosineSimilarity}(M_{ct}(C), M_{cv}(F))$
        \ForAll{sub-sentence $c_j \in C$}
            \If{$\text{Average}(\text{TopK}(S[j,:], K=3)) < \alpha$}
                \State \Return (False, \{problematic sub-sentences\})
            \EndIf
        \EndFor
        \State \Return (True, $\emptyset$)
    \EndProcedure

    \Statex
    \Procedure{GenerateAndVerify}{Prompt $p$, Model ${M}$, $\alpha$, $N_{retry}$, context=None}
        \For{$i = 1 \to N_{retry}$}
            \State $o \gets {M}(p, \text{context})$
            \State is\_consistent, issues $\gets \textsc{Verify}(\mathcal{V}, o, \alpha)$
            \If{is\_consistent}
                \State \Return $o$
            \Else
                \State $p \gets \text{CreateRefinementPrompt}(p, \text{issues})$
            \EndIf
        \EndFor
        \State \Return $\text{HumanReviewQueue.add}(\mathcal{V}, o)$
    \EndProcedure
\end{algorithmic}
\end{algorithm}

\subsection{Computational Cost of the Offline CoE Text Generation Pipeline}
\label{sec:coe_cost}
As a supplement to the main paper, we report the detailed per-stage computational cost of the offline CoE text generation pipeline (Algorithm~\ref{alg:cot_pipeline_separated}) here. Note that this pipeline is executed only once, offline, on the support set before model training; once the generated texts are cached as CLIP features, they introduce no additional overhead during model training or test-time localization. As shown in Table~\ref{tab:coe_cost_sup}, the full pipeline takes about 34.3 seconds per support video on average, measured on 4$\times$NVIDIA RTX A6000 GPUs.

\begin{table}[h]
\centering
\caption{Computational cost of the offline CoE text generation pipeline per support video. The per-stage time already incorporates the average of 2.4 refinement rounds per video.}
\label{tab:coe_cost_sup}
\begin{tabular}{@{}lc@{}}
\toprule
\textbf{Stage} & \textbf{Avg. Cost / Video} \\ \midrule
Stage 1: VLM detailed description & 4.7 s \\
Stage 2: VLM key-event filtering  & 3.4 s \\
Stage 3: LLM CoE reasoning        & 5.3 s \\
CLIP-based verification           & 0.9 s \\
Avg. \# refinement iterations     & 2.4 \\ \midrule
\textbf{Total per-video cost}     & \textbf{34.3 s} \\ \bottomrule
\end{tabular}
\end{table}

\section{Compared Methods.}
To validate the effectiveness of our proposed approach, we compare it with the following state-of-the-art methods: \textbf{1)} Feng~\emph{et al}~\cite{feng2018video} propose a cross-gated bilinear matching model; \textbf{2)} Hu~\emph{et al}~\cite{hu2019silco} introduce a spatial similarity module combined with feature reweighting via graph convolutional networks; \textbf{3)} Yang~\emph{et al}~\cite{yang2020localizing} design a progressive alignment network to iteratively fuse support information into the query branch; \textbf{4)} Yang~\emph{et al}~\cite{yang2021few} propose a few-shot transformer architecture optimized for joint commonality learning and localization without proposals; \textbf{5)} Nag~\emph{et al}~\cite{nag2021few} introduce a query-adaptive Transformer capable of adapting to new classes; \textbf{6)} Hsieh~\emph{et al}~\cite{hsieh2023aggregating} propose an Aggregating Bilateral Attention mechanism that uncovers query-support dependencies through embedding norms and context awareness; \textbf{7)} Lee~\emph{et al}~\cite{lee2023few} develop a cross-correlation attention mechanism to transform support features into the query context; \textbf{8)} Zhang~\emph{et al}~\cite{CGCN2025} propose the Context Graph Convolutional Network and utilize the multi-scale graph convolutions to model sequence, intra-action, and inter-action relationships.

\section{More details about the HAL dataset}
\label{More Details about HAL Dataset}

\subsection{More Information about the Anomaly Types} 
In our HAL dataset, there are a total of 12 human-related anomaly categories, which are as follows: Animal Attack, Assault, Abuse, Fighting, People Falling, Public Safety, Pedestrian Incidents, Robbery, Riot, Shooting, Social Unrest, Theft, and Vandalism. The specific quantities and the proportions are shown in Table \ref{tab:anomaly_counts}.

\begin{table}[h]
    \centering
    \caption{The number of videos for each anomaly type based on the provided data.}
    \setlength{\tabcolsep}{5mm}
    \label{tab:anomaly_counts}
    \begin{tabular}{l c}
        \toprule
        \textbf{Anomaly Type} & \textbf{Number} \\
        \midrule
        Robbery             & 235 \\
        Fighting            & 211 \\
        Shooting            & 153 \\
        Riot                & 129 \\
        Vandalism           & 84 \\
        Theft               & 83 \\
        Animal Attack       & 56 \\
        People Falling      & 43 \\
        Pedestrian Incidents& 40 \\
        Assault             & 20 \\
        Abuse               & 11 \\
        Setting Fire        & 7 \\ 
        \midrule
        \textbf{Total}      & \textbf{1072} \\
        \bottomrule
    \end{tabular}
\end{table}

\subsection{More visual examples of the HAL dataset} 
\begin{figure*}[h]
    \centering
    \includegraphics[width=1\linewidth]{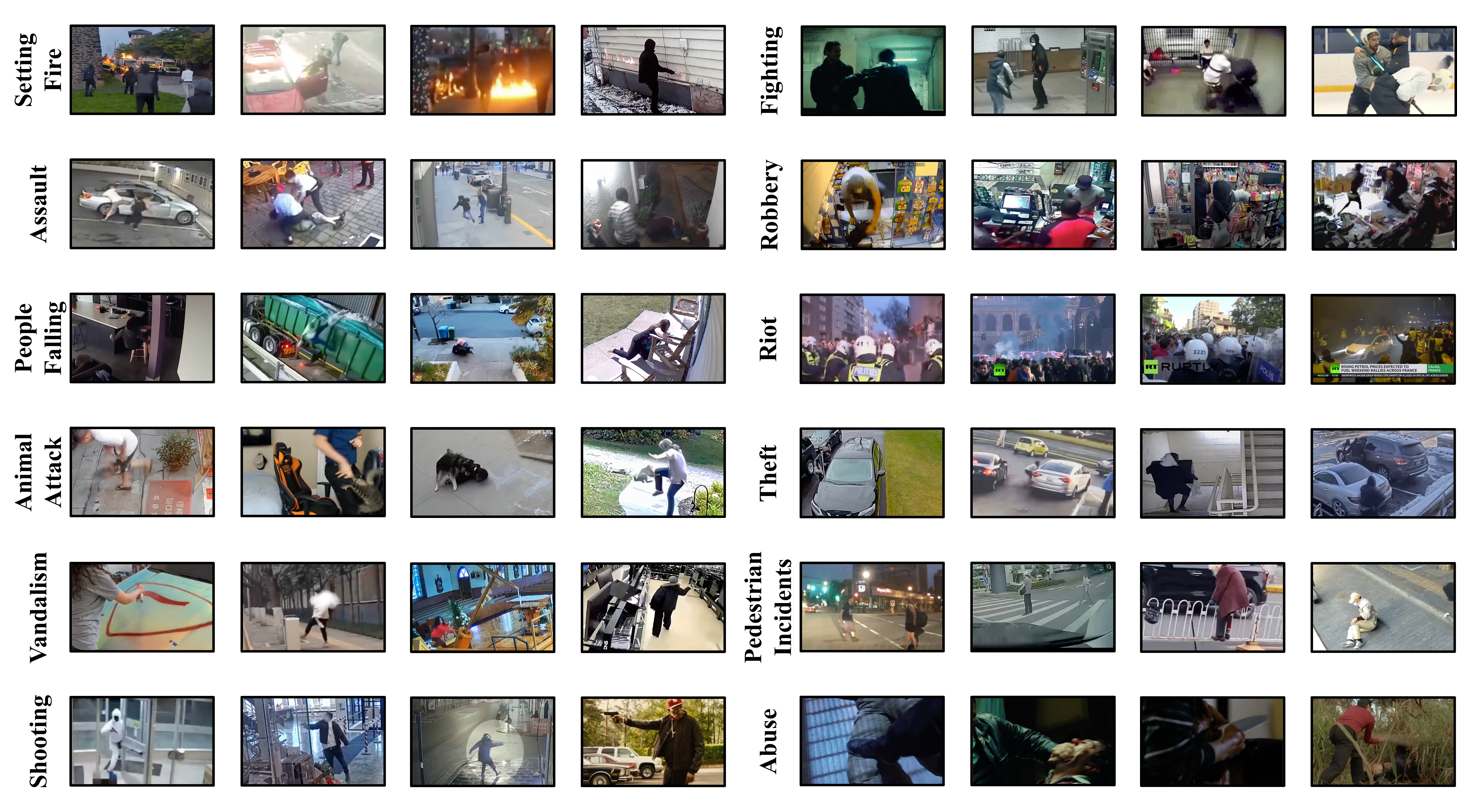}
    \caption{The video samples include instances of different categories.}
    \label{fig:sample1}
\end{figure*}

The HAL dataset features a diverse range of scene types, including stores, streets, private residences, trains, and more. Additionally, the video formats are varied, encompassing surveillance footage, news reports, short videos, and cinematic scenes. In Figure \ref{fig:sample1}, we present more visuals to facilitate the observation of abnormal events across different scenes and video types. This diversity not only enhances the richness of the dataset but also provides a solid training foundation for the model's generalization capabilities. By training the model with abnormal events in different scenes, we hope to further improve the model's ability to detect anomalous behaviours in complex environments. 

\subsection{More CoE text examples of the HAL dataset} 
\begin{figure*}[h]
    \centering
    \includegraphics[width=1.0\linewidth]{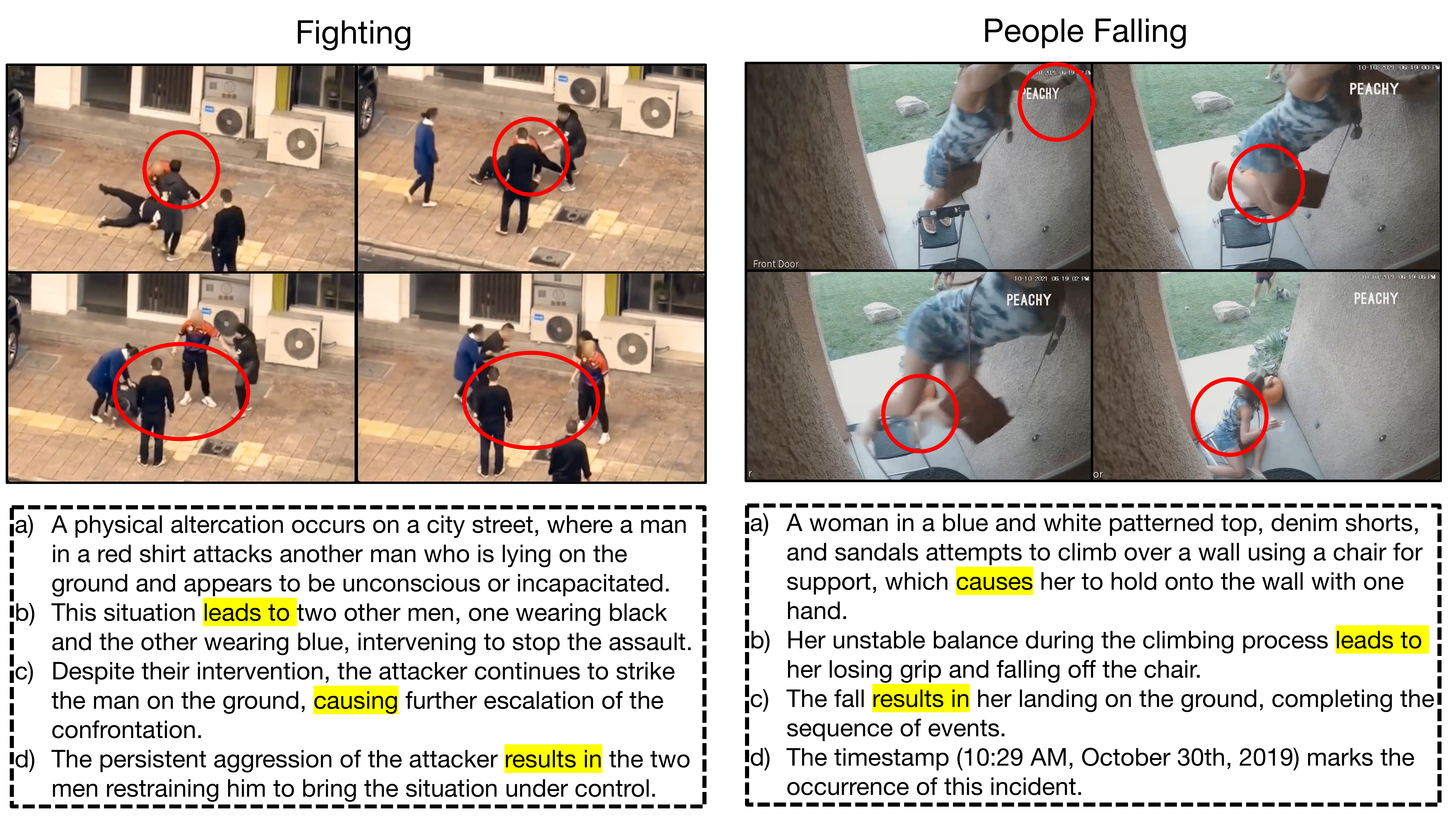}
    \caption{The CoE text of ``Fighting" and ``People Falling" anomaly type. }
    \label{fig:text_1}
\end{figure*}

\begin{figure*}[h]
    \centering
    \includegraphics[width=1.0\linewidth]{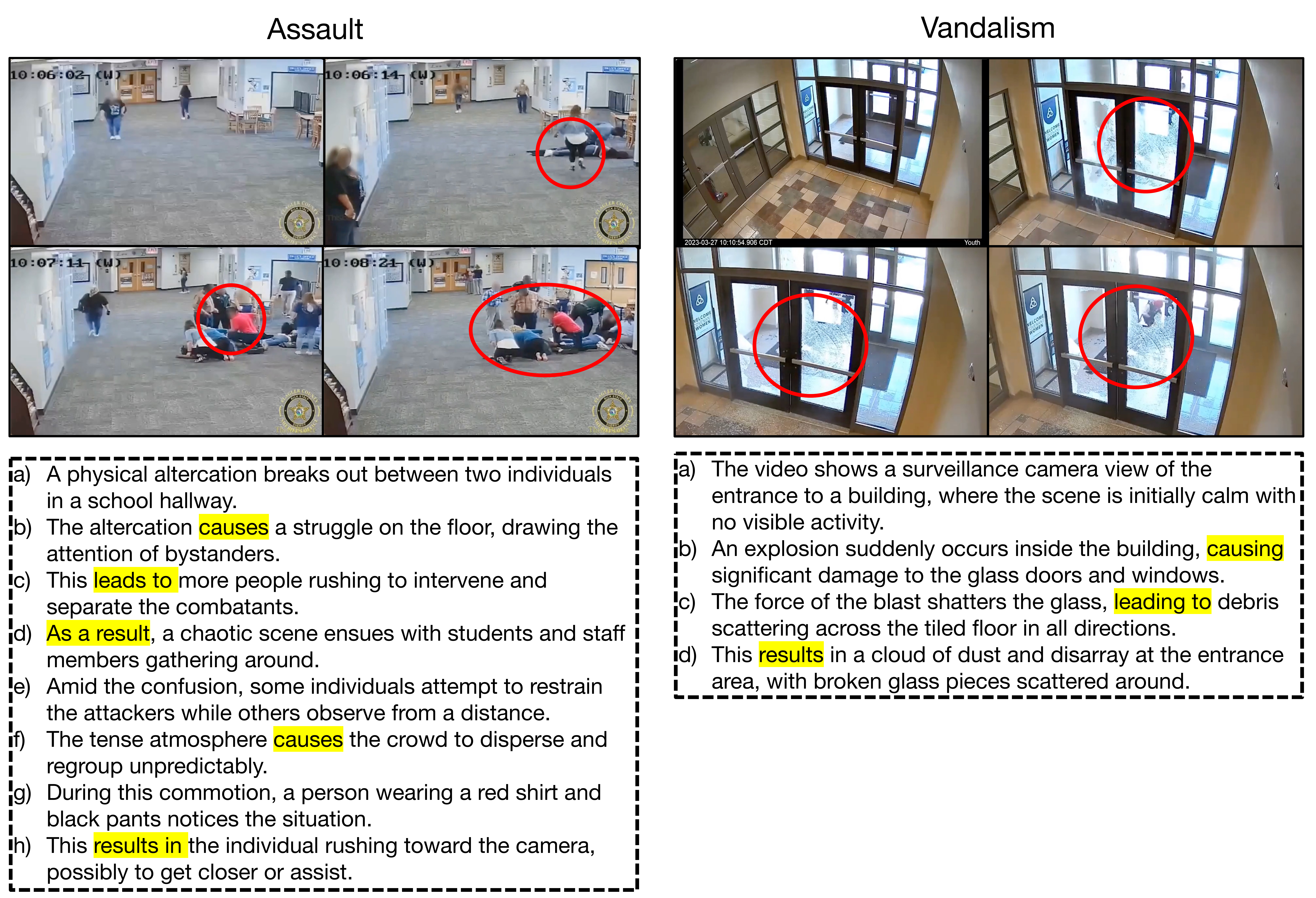}
    \caption{The CoE text of ``Assault" and ``Vandalism" anomaly type.}
    \label{fig:text_3}
\end{figure*}

We not only generated frame-level descriptions but also more comprehensive CoE text for the HAL dataset. We selected six types of anomalies from the dataset and presented their corresponding CoE text in Figure \ref{fig:text_1} $\sim$ Figure \ref{fig:text_3}. 
Compared to captions, CoE text can more comprehensively express the complete process of anomalous events and provide more comprehensive textual guidance, thereby enhancing the accuracy of action localization.

\begin{table}[t]
\centering
\caption{
    Human evaluation of CoT text quality, with a lower average rank indicating higher quality and preference.
}
\label{tab:user_study}
\setlength{\tabcolsep}{10pt}
\renewcommand{\arraystretch}{1.1}
\begin{tabular}{lc}
\toprule
\textbf{Method} & \textbf{Average Rank} $\downarrow$ \\
\midrule
Qwen2.5-VL-72B-AWQ & 2.27 \\
Qwen2.5-VL-7B + Qwen3-30B & 2.09 \\
\textbf{Ours (VideoChat + DeepSeek-R1)} & \textbf{1.64} \\
\bottomrule
\end{tabular}
\end{table}
\subsection{Human evaluation of the CoE text quality}

Beyond factual accuracy, the core advantage of CoE text lies in its logical coherence and inferential completeness. 
To quantitatively evaluate the quality of our generated text in these aspects, we conducted a human evaluation study. 
In this study, we compared the outputs of our method (VideoChat + DeepSeek-R1) against text generated by two baseline approaches: a powerful standalone VLM (Qwen2.5-VL-72B-AWQ) and a VLM+LLM combination (Qwen2.5-VL-7B + Qwen3-30B).

Our evaluation protocol involved 10 human participants who were asked to rank the outputs from the three methods based on four key criteria: 
1) \textbf{Logical Soundness}: Is each step of the reasoning logical? 
2) \textbf{Factuality}: Is the reasoning grounded in facts from the video? 
3) \textbf{Completeness}: Are any key reasoning steps missing? 
4) \textbf{Conciseness}: Are there unnecessary or redundant reasoning steps?

The results, summarized in Table~\ref{tab:user_study}, clearly demonstrate the superiority of our method. 
In the side-by-side ranking, the CoE text generated by our approach was strongly preferred by human evaluators, achieving a superior average rank of 1.64. 
This significantly outperforms the baselines, which scored 2.09 and 2.27, respectively. 
This indicates that our stage-wise reasoning pipeline not only produces factually consistent text but also generates higher-quality output in terms of logic, completeness, and conciseness, providing a superior semantic guide for the subsequent localization task.

\section{Additional ablation and visualization analysis}
\subsection{Impact of the number of the STPE blocks}
We analyze the impact of the number of STPE blocks on action localization performance. We fix the number of semantic nodes at 6 and the number of semantic pyramid layers at 3, conducting experiments under the multi-instance setting of THUMOS14, with results shown in Table~\ref{tab:ablation_stpe_blocks}. Due to the limited number of video samples in the THUMOS14 dataset, the model cannot be adequately trained when the number of blocks exceeds 2, resulting in a decline in performance.
\begin{table}[]
\centering
\caption{
    Ablation study of the number of STPE blocks.
}
\setlength{\tabcolsep}{3mm}
\label{tab:ablation_stpe_blocks}
\renewcommand{\arraystretch}{1.1}
\begin{tabular}{lccccc}
\toprule
\textbf{Number of Blocks} & 1 & 2 & 3 & 4 & 5 \\
\midrule
\textbf{mAP@0.5} & 14.5 & \textbf{18.2} & 16.6 & 13.2 & 11.9 \\
\bottomrule
\end{tabular}
\end{table}

\subsection{Visualization Analysis}
\begin{figure*}[h]
    \centering
    \includegraphics[width=0.95\linewidth]{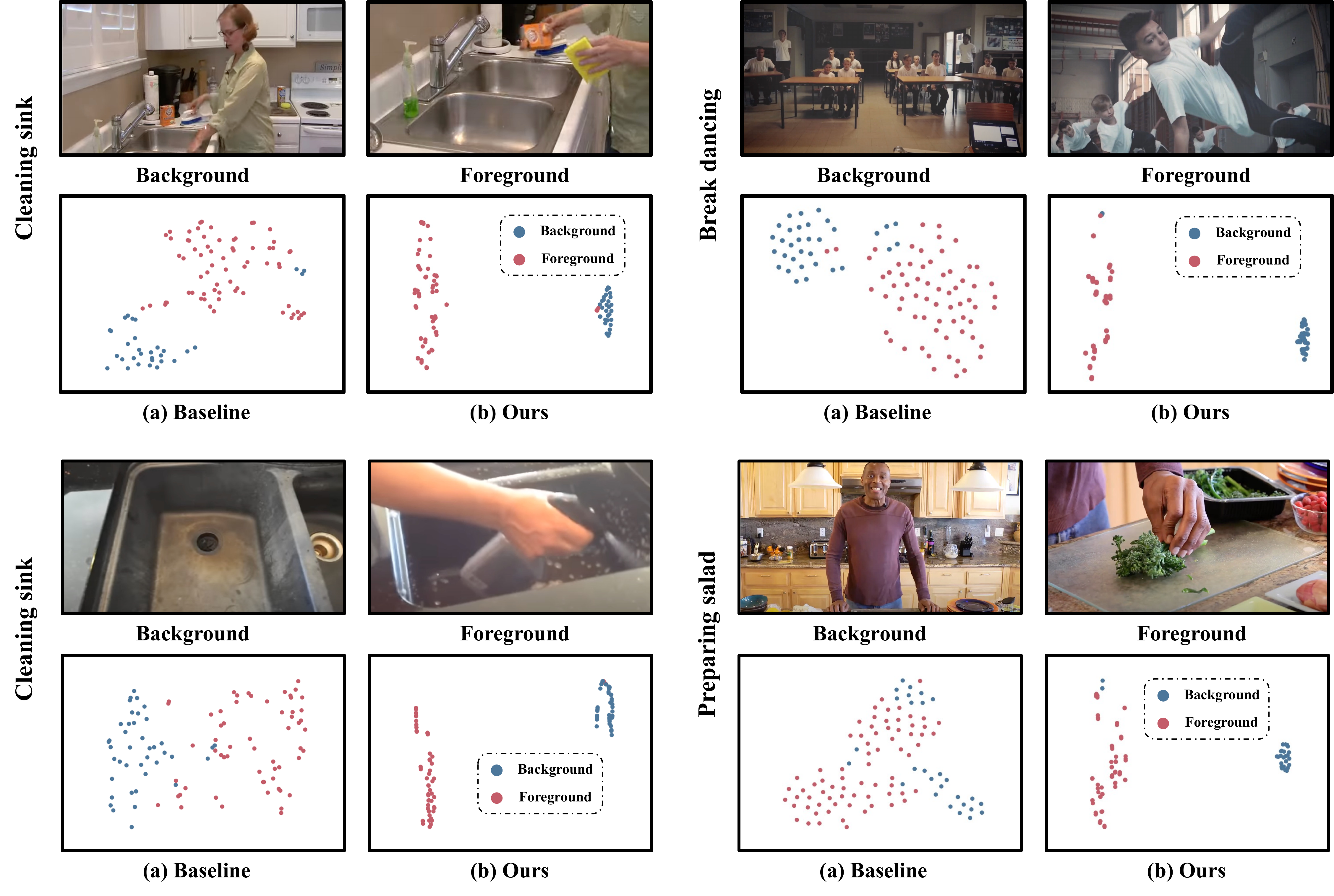}
    \caption{T-SNE visualization of foreground and background features on “Cleaning sink”, ``Break dancing" and ``Preparing salad" of ActivityNet1.3 dataset. Red points and
blue points denote foreground and background, respectively.
}
    \label{feature_visualization}
\end{figure*}

\begin{figure*}[h]
    \centering
    \includegraphics[width=1.04\linewidth]{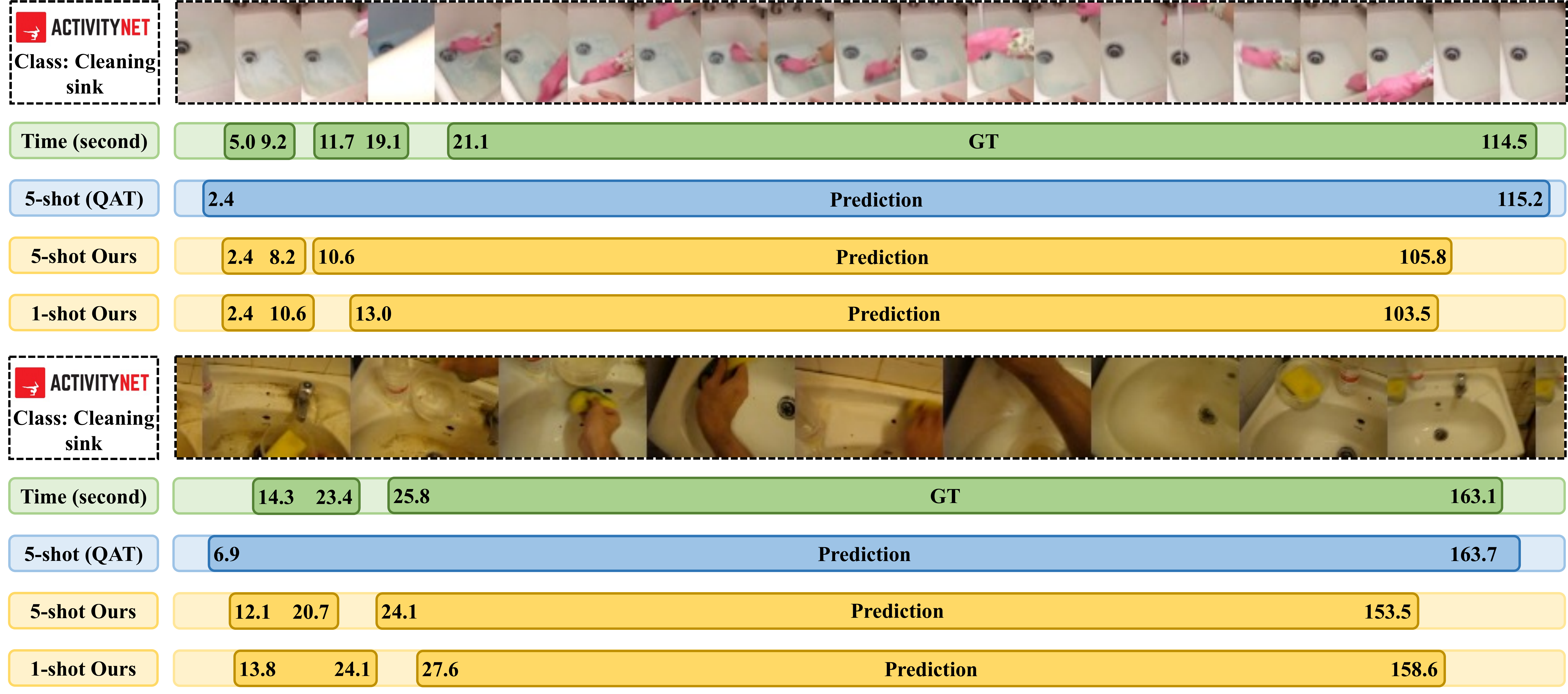}
    \caption{Qualitative comparisons of our method (1-shot and 5-shot setting), QAT~\cite{nag2021few} (5-shot setting) and Ground Truth (GT) on “Cleaning sink~(multi-instance) of ActivityNet1.3 dataset.}
    \label{qualitative_result_4}
\end{figure*}

\begin{figure*}[h]
    \centering
    \includegraphics[width=1.04\linewidth]{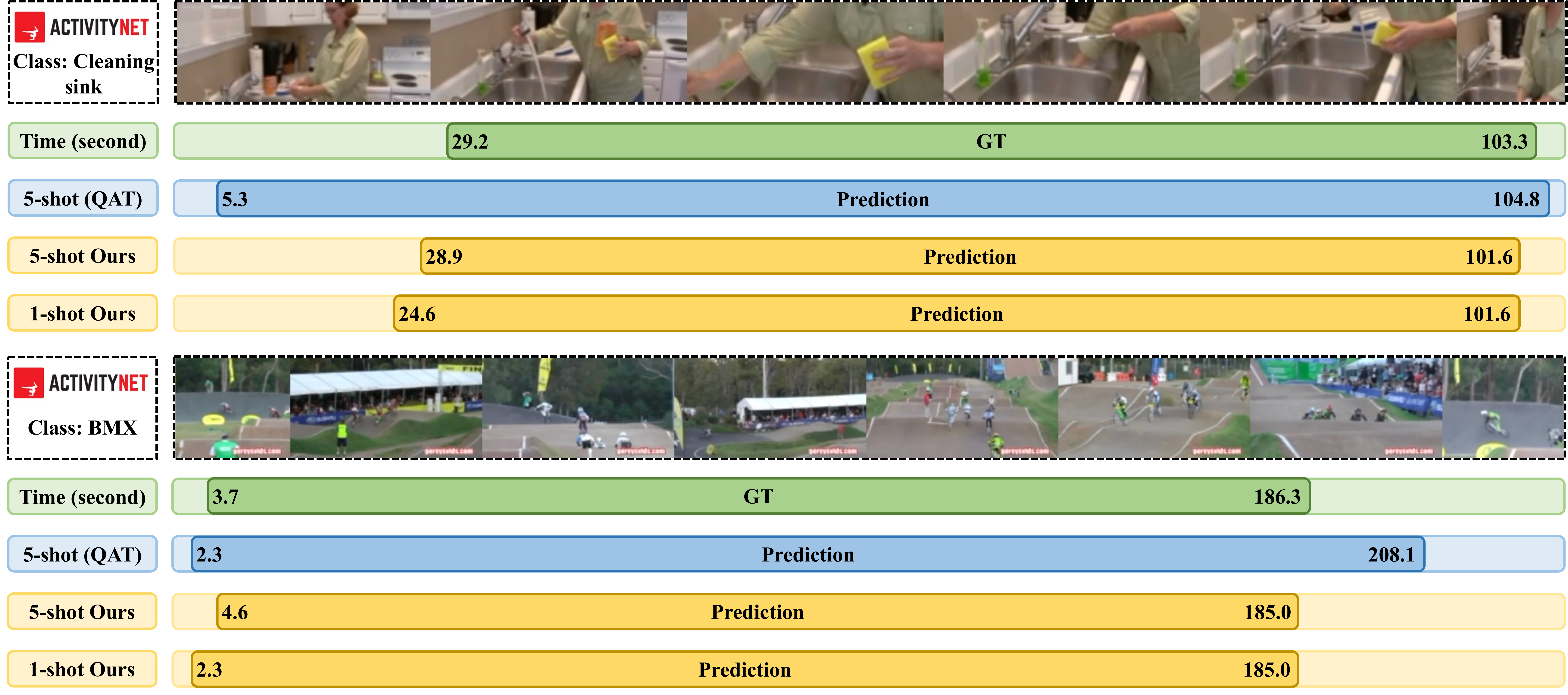}
    \caption{Qualitative comparisons of our method (1-shot and 5-shot setting), QAT~\cite{nag2021few} (5-shot setting) and Ground Truth (GT) on “Cleaning sink"~(single-instance) and “BMX"~(single-instance) of ActivityNet1.3 dataset.}
    \label{qualitative_result_6}
\end{figure*}

\begin{figure*}[h]
    \centering
    \includegraphics[width=1.04\linewidth]{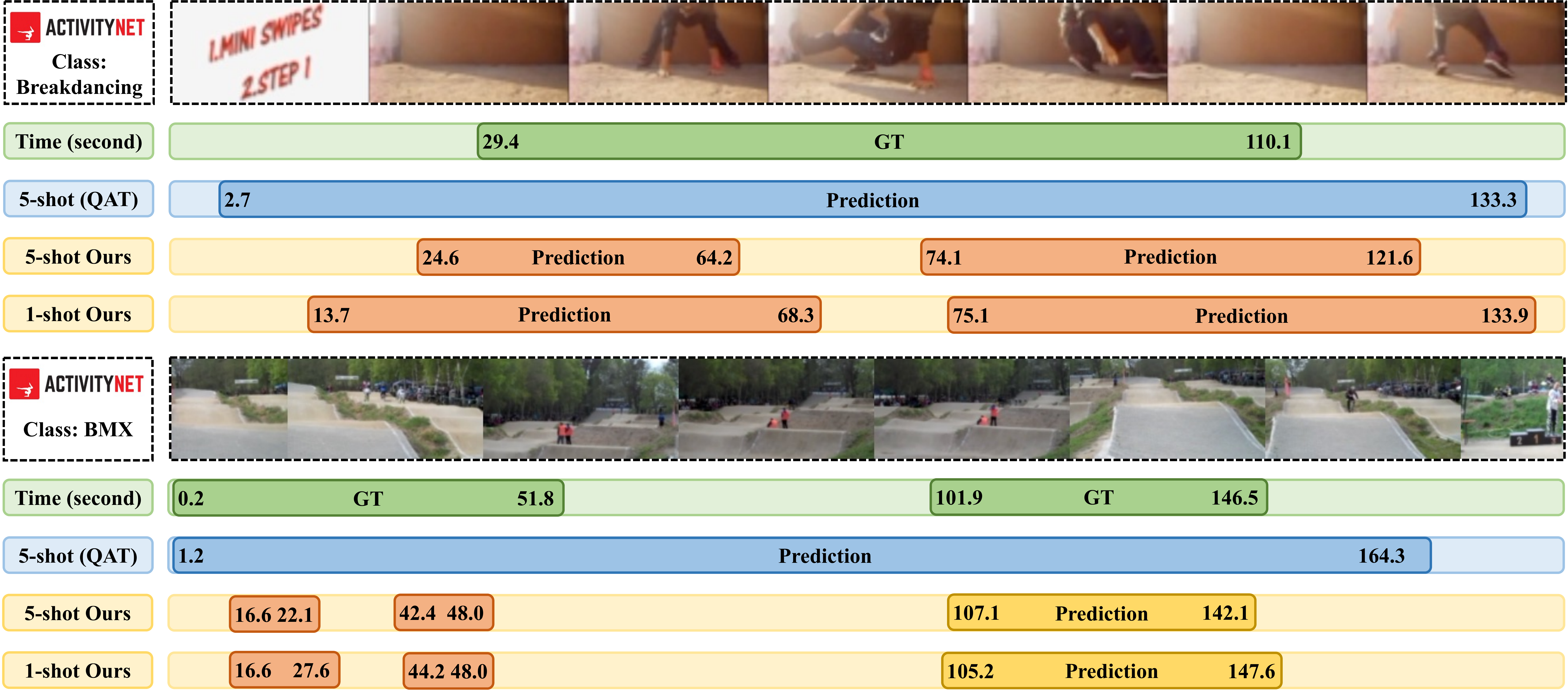}
    \caption{Qualitative comparisons of our method (1-shot and 5-shot setting), QAT~\cite{nag2021few} (5-shot setting) and Ground Truth (GT) on “Breakdancing"~(single-instance) and “BMX"~(multi-instance) of ActivityNet1.3 dataset.}
    \label{qualitative_result_3}
\end{figure*}

\begin{figure*}[h]
    \centering
    \includegraphics[width=1.04\linewidth]{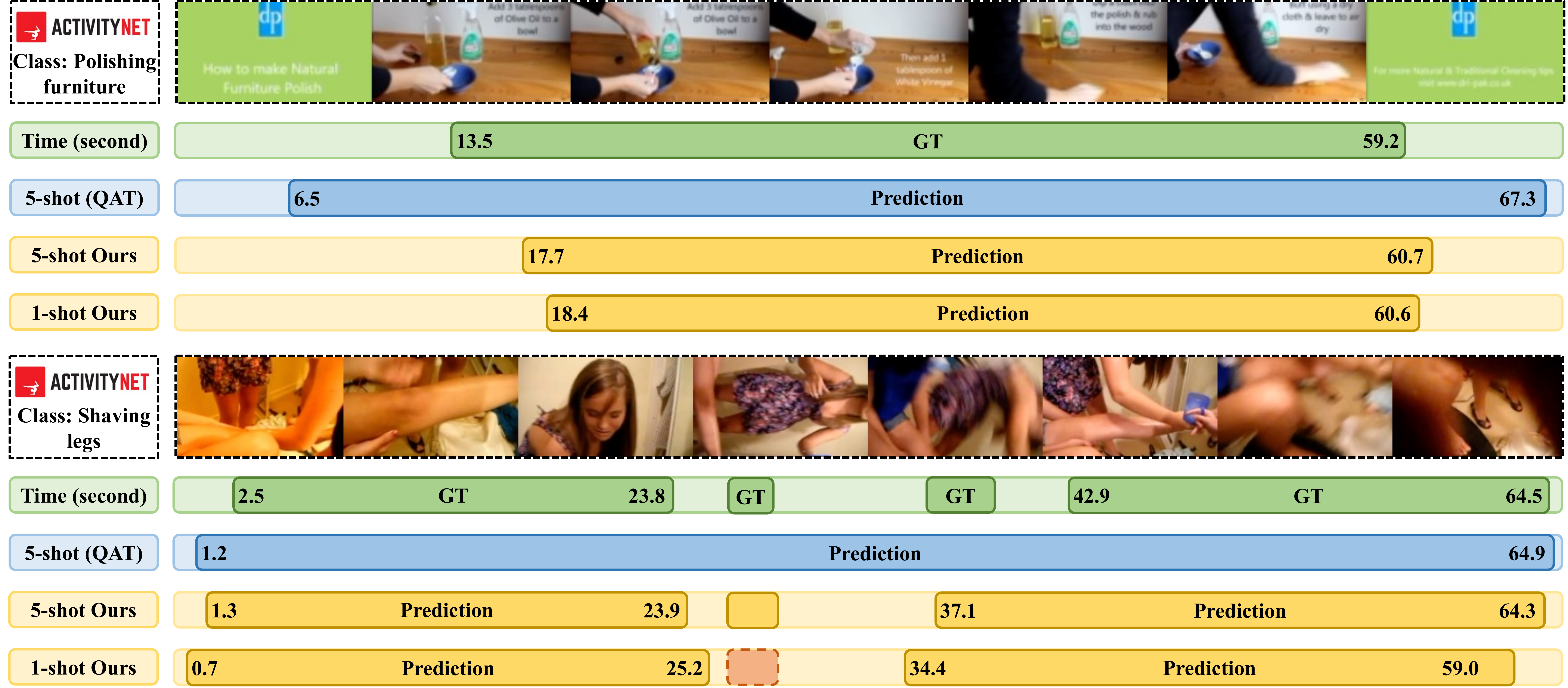}
    \caption{Qualitative comparisons of our method (1-shot and 5-shot setting), QAT~\cite{nag2021few} (5-shot setting) and Ground Truth (GT) on “Rope skipping"~(single-instance) and “Shaving legs"~(multi-instance) of ActivityNet1.3 dataset.}
    \label{qualitative_result_2}
\end{figure*}

To better illustrate the effectiveness of our method, we utilized T-SNE to visualize a sample from ActivityNet1.3 in Figure~\ref{feature_visualization}. The left side represents the original C3D features, while the right side shows the features after incorporating the STPE. Besides,
we also select two sets of single-instance and multi-instance results from the ActivityNet1.3 dataset for visualization, one including correctly localized instances and the other containing incorrectly localized instances. We observe that in both sets of samples, QAT only achieves relatively coarse localization, whereas our proposed method provides more accurate localization results, whether for 1-shot or 5-shot. Additionally, as shown in Figure~\ref{qualitative_result_4}$\sim$ Figure~\ref{qualitative_result_6},  our method achieves more consistent localization proposals under both multi-instance and single-instance scenarios, even when there are numerous short regions. In Figure~\ref{qualitative_result_3} and Figure~\ref{qualitative_result_2}, we show several samples with localization errors (red background box). In these cases, our method incorrectly classifies some background snippets as foreground snippets. This may be due to the presence of similar foreground and background snippets in the original C3D snippet features or overly subtle action performers in ActivityNet1.3 videos. It leads to erroneous semantic attention operations in STPE, pushing these background snippets closer to the foreground snippets. Consequently, these background snippets are mistakenly localized as action regions during the post-processing phase.

\begin{figure*}[h]
    \centering
    \includegraphics[width=1.0\linewidth]{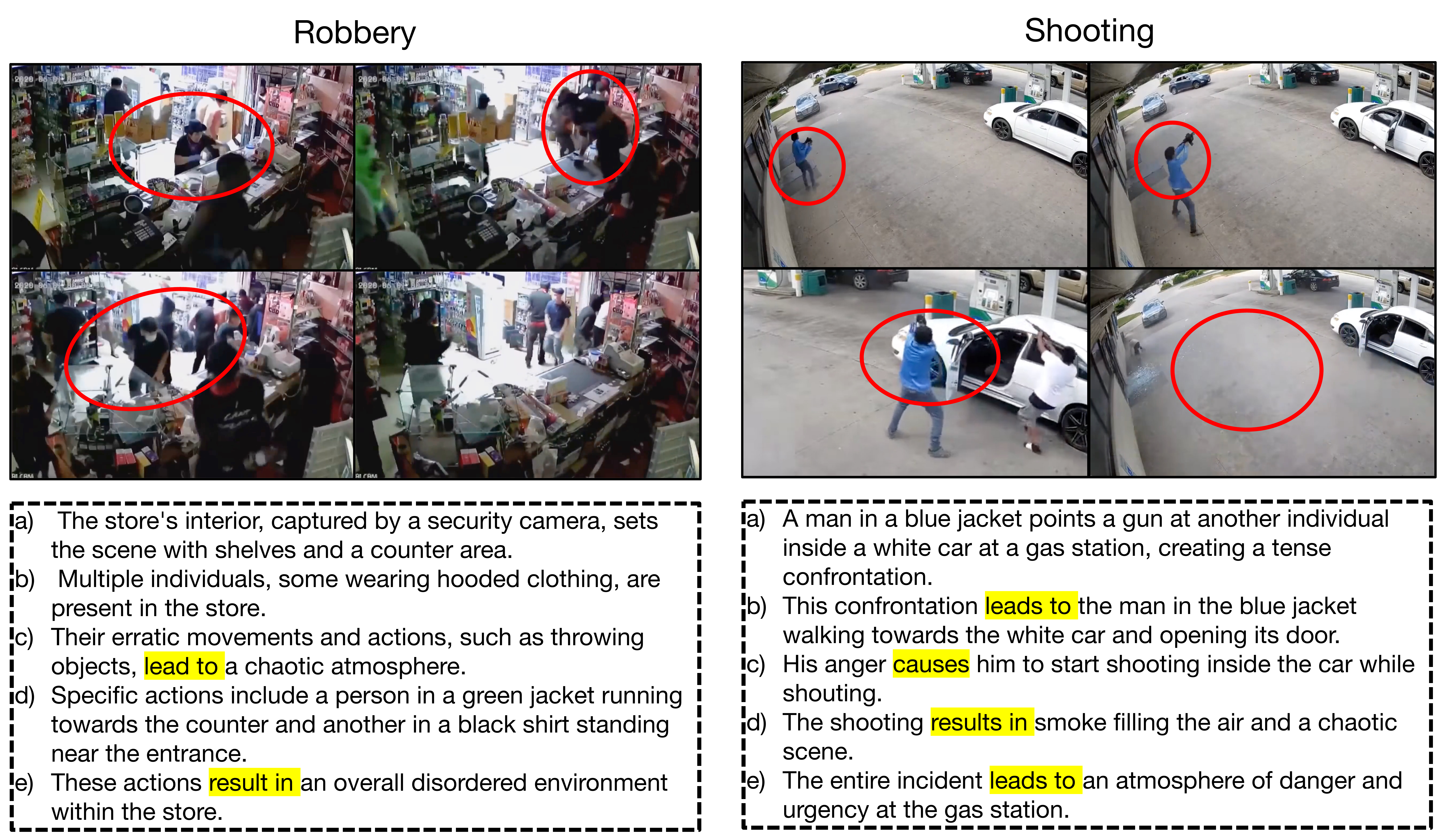}
    \caption{The CoE text of ``Robbery" and ``Shooting" anomaly type.}
    \label{fig:text_2}
\end{figure*}

\begin{figure*}[h]
    \centering
    \includegraphics[width=1.04\linewidth]{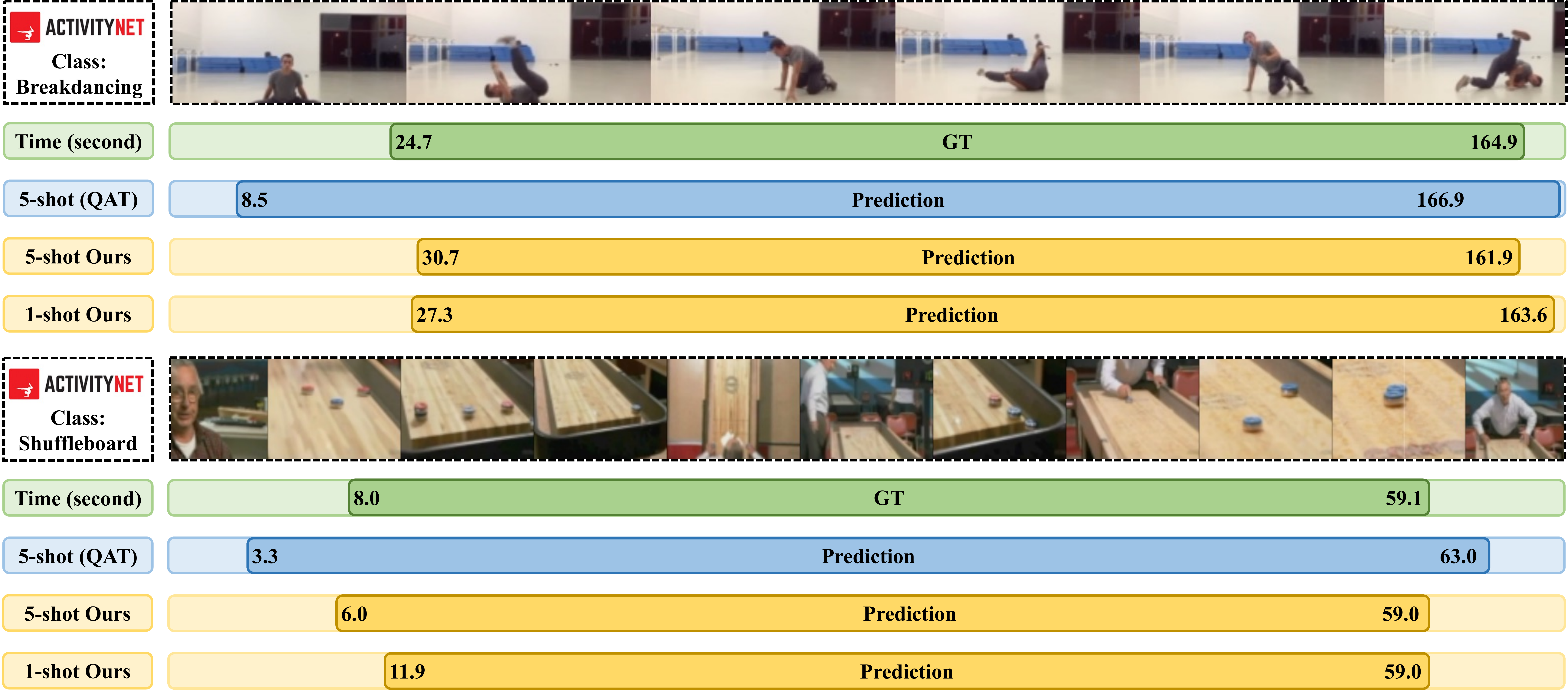}
    \caption{Qualitative comparisons of our method (1-shot and 5-shot setting), QAT~\cite{nag2021few} (5-shot setting) and Ground Truth (GT) on “Breakdancing"~(single-instance) and “Shuffleboard"~(single-instance) of ActivityNet1.3 dataset.}
    \label{qualitative_result_5}
\end{figure*}

